%% file: neurips_2024.tex
\documentclass{article}

\usepackage[preprint]{neurips_2024}

\usepackage[utf8]{inputenc} 
\usepackage[T1]{fontenc}    
\usepackage{hyperref}       
\usepackage{url}            
\usepackage{booktabs}       
\usepackage{amsfonts}       
\usepackage{nicefrac}       
\usepackage{microtype}      
\usepackage{xcolor}         
\usepackage{bm}

\usepackage{amsmath}
\usepackage{wrapfig}
\usepackage{graphicx}

\usepackage{algorithm}
\usepackage{algorithmic}
\usepackage{xcolor}
\usepackage{makecell}
\usepackage{multirow}
\usepackage{subcaption}
\newcommand{\revise}[2]{#2}
\newcommand{\cameraready}[1]{#1}
\newcommand{\submission}[1]{}

\definecolor{mydarkblue}{rgb}{0,0.08,0.8}
\hypersetup{colorlinks,linkcolor={mydarkblue},citecolor={mydarkblue},urlcolor={red}}

\title{Hindsight Preference Learning for Offline Preference-based Reinforcement Learning}

\author{%
    Chen-Xiao Gao$^{1,2}$, Shengjun Fang$^{1,2}$, Chenjun Xiao$^3$, Yang Yu$^{1,2}$, Zongzhang Zhang$^{1,2}$\thanks{Correspondence to: Zongzhang Zhang <zzzhang@nju.edu.cn>.}\\
    \\
    \small{
    $^1$National Key Laboratory for Novel Software Technology, Nanjing University, Nanjing, China}\\
    \small{$^2$School of Artificial Intelligence, Nanjing University, Nanjing, China}\\
    \small{$^3$The Chinese University of Hong Kong, Shenzhen, China
    }\\
    \texttt{\{gaocx,fangsj\}@lamda.nju.edu.cn},\ \texttt{chenjunx@cuhk.edu.cn}\\
    \texttt{\{yuy,zzzhang\}@nju.edu.cn}
}

\begin{document}

\maketitle

\begin{abstract}
\input{tex/abstract.tex}
\end{abstract}

\section{Introduction}
\input{tex/1.introduction.tex}

\vspace{-2mm}
\section{Preliminaries}
\input{tex/3.preliminary.tex}

\vspace{-2mm}
\section{Hindsight Preference Learning}

\input{tex/4.method.tex}

\vspace{-2mm}
\section{Related Work}
\input{tex/2.related_work.tex}

\vspace{-2mm}
\section{Experiments}\label{sec:experiments}
\input{tex/5.experiments.tex}

\vspace{-2mm}
\section{Conclusions and Discussions}
\input{tex/6.conclusions.tex}

\bibliography{neurips_2024}
\bibliographystyle{plainnat}
\newpage
\appendix
\input{tex/appendix.tex}

\submission{
\newpage
\input{tex/checklist.tex}
}

\end{document}

%% file: tex/abstract.tex

Offline preference-based reinforcement learning (RL), which focuses on optimizing policies using human preferences between pairs of trajectory segments selected from an offline dataset, has emerged as a practical avenue for RL applications. Existing works rely on extracting \emph{step-wise} reward signals from \emph{trajectory-wise} preference annotations, assuming that preferences correlate with the cumulative Markovian rewards. However, such methods fail to capture the holistic perspective of data annotation: Humans often assess the desirability of a sequence of actions by considering the overall outcome rather than the immediate rewards. To address this challenge, we propose to model human preferences using rewards conditioned on future outcomes of the trajectory segments, i.e. the \textit{hindsight information}. For downstream RL optimization, the reward of each step is calculated by marginalizing over possible future outcomes, the distribution of which is approximated by a variational auto-encoder trained using the offline dataset. Our proposed method, \emph{Hindsight Preference Learning (HPL)}, can facilitate credit assignment by taking full advantage of vast trajectory data available in massive unlabeled datasets. Comprehensive empirical studies demonstrate the benefits of HPL in delivering robust and advantageous rewards across various domains. \cameraready{Our code is publicly released at \url{https://github.com/typoverflow/WiseRL}.}

%% file: tex/1.introduction.tex

Although reinforcement learning (RL) has demonstrated remarkable success in various sequential decision-making tasks~\citep{alphastar, hok}, its application in real-world scenarios remains challenging for practitioners, primarily due to two key reasons. First, modern RL methods typically require millions of online interactions with the environment~\citep{sac}, which is prohibitively expensive and potentially dangerous in embodied applications~\citep{levine_survey}. Second, reward engineering is necessary to align the induced behavior of the policy with human interests~\citep{reward_engineer1,reward_engineer2}. 
However, tweaking the reward requires substantial efforts and extensive task knowledge of the real-world scenarios.  
Reward hacking frequently occurs when the reward is improperly configured, leading to unintended consequences~\citep{reward_hacking}.


There are several research directions aiming for addressing above challenges~\citep{reward_learning1,reward_learning2,reward_learning3}, among which \emph{offline Preference-based RL (offline PbRL)} has gained increasing attention recently~\citep{ipl,dppo,oppo}. 
In offline PbRL, an offline dataset is collected by deploying a behavior policy, after which human labelers are  required to provide relative preferences between two trajectory segments selected from the offline dataset. 
Offline PbRL significantly reduces the burden on human effort given that labeling preferences between trajectories is considerably easier compared to crafting step-wise reward signals. 
It has proven effective in large-scale applications including fine-tuning large language models~\citep{instruct_gpt, llama2, dpo}.


\begin{figure*}[t]
    \centering
    \includegraphics[width=0.86\linewidth]{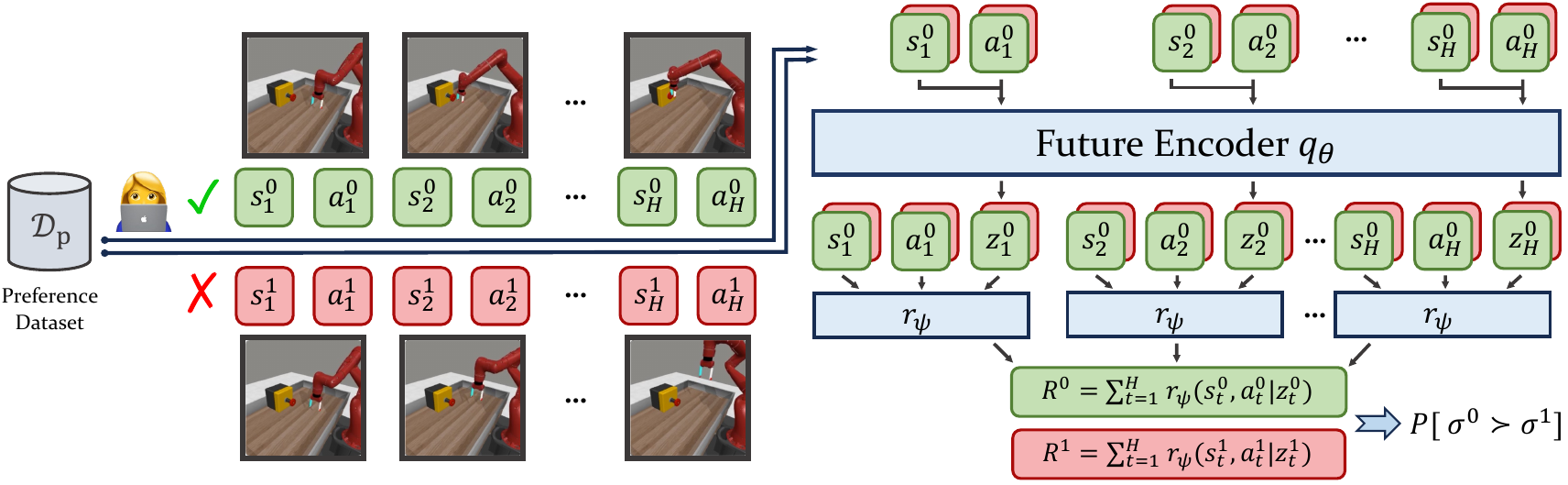}
    \caption{Illustration of the reward learning procedure in HPL. Unlike previous methods, HPL first generates embeddings $z_t$ to encode the future part of the segments and optimize a reward function $r_{\psi}$ which is conditioned on the $s_t$, $a_t$ and the future $z_t$ using the Bradley-Terry model. }
    \label{fig:hpm}
\end{figure*}

Offline PbRL typically follows a two-phase paradigm: 1) learning a reward function that aligns with human preferences using a small labeled {preference dataset}; 2) applying the reward function to label a massive {unlabeled dataset} and performing policy optimization~\citep{christiano1}. 
For the first phase, existing methods employ Bradley-Terry model~\citep{btmodel} to learn \emph{step-wise} reward signals from \emph{trajectory-wise} preferences based on the \emph{Markovian reward assumption}: the preference correlates with the {cumulative rewards} of each trajectory~\citep{christiano1}. 
However, as previous works~\citep{pt} have unveiled, such an assumption is technically flawed and limited since humans evaluate the trajectory segments from a global perspective, making the obtained reward bear an implicit dependence on the future part of the segment. Consider the case of purchasing lottery tickets as an example. Although the expected return is negative, the payoff can be substantial if one wins. However, when assigning credits, we should allocate higher rewards to purchasing tickets only when we are certain of winning, rather than unconditionally encouraging such behavior.


In light of this, we propose \emph{Hindsight Preference Learning (HPL)} to account for such dependence on future information. 
The key idea of HPL is to develop a \emph{hindsight preference model}, which models human preferences using a reward function conditioning on the state $s$, action $a$ and the future trajectory starting from $(s, a)$, i.e. the \textit{hindsight information}. 
In particular, given a $H$-length trajectory $\sigma_{1:H}=(s_1, a_1, s_2, a_2, \dots, s_H, a_H)$, the reward of $s_t, a_t$ for $1\leq t\leq H$ is given by  $r(s_t, a_t | \sigma_{t:t+k})$. When labeling the unlabeled dataset, a scalar reward signal is computed for each state-action pair by marginalizing over all possible hindsight information, $r(s,a) = \int_{\sigma} p(\sigma | s, a) r(s,a | \sigma) d \sigma$. 
To deal with the high-dimensional nature of hindsight information, we pre-train a variational auto-encoder to efficiently represent the hindsight information, making the above marginalization feasible in practice. The reward learning procedure of HPL is illustrated in Figure~\ref{fig:hpm}.   

HPL has two key benefits over prior works. First, by leveraging hindsight information in preference modeling, it captures the implicit holistic perspective of human preference labeling, addressing the key limitation of the Markovian reward assumption adopted in previous works.   
Second, the two-phase paradigm of offline PbRL might become less effective if there is a substantial distribution mismatch between preference and unlabeled dataset. HPL can take advantage of the unlabeled dataset by learning a prior over future outcomes. This allows for incorporating the information of trajectory distribution carried by the unlabeled dataset, thus delivering robust and advantageous rewards when labeling the unlabeled dataset. 
We provide extensive evaluations to verify these benefits.


%% file: tex/3.preliminary.tex
\subsection{Markov Decision Process}
In standard RL, an agent interacts with an environment characterized by a Markov Decision Process (MDP) $\langle\mathcal{S}, \mathcal{A}, r, T, \gamma\rangle$. Here, $\mathcal{S}$ and $\mathcal{A}$ represent the state space and action space respectively, while $r(s, a)$ denotes the reward function, $T(s'|s, a)$ is the transition function, and $\gamma$ is the discounting factor. 
The value function defines the expected cumulative reward by following a policy $\pi(a|s)$,
\begin{align}
    V^\pi(s) = \mathbb{E} \left[\sum_{t=0}^\infty \gamma^tr(s_t, a_t) | s_0 = s \right] \, .
\end{align}
The primary objective of RL algorithms is to find an optimal policy that maximizes  $\mathbb{E}_{s_0\sim\mu_0} [V^\pi(s_0)]$, where $\mu_0$ is the initial state distribution. 


\subsection{Offline Preference-Based Reinforcement Learning}\label{sec:preliminary_oprl}

In this work, we consider the problem of learning an optimal decision-making policy from a previously collected dataset with \emph{preference feedback}. 
In its generalist framework, the ground-truth reward is not given in the data. Instead, the learner is provided with a \emph{preference dataset} and a massive \emph{unlabeled dataset}, and follows a two-phase paradigm: 1) \emph{reward learning},  learn a reward model $r$ with the  preference data;
and 
2) \emph{reward labeling}, apply $r$ to label the unlabeled dataset in order to perform policy optimization with a large amount of data. 

Let $\sigma=(s_0, a_0, s_1, a_1,\dots, s_{|\sigma|}, a_{|\sigma|})$ denote a consecutive segment of states and actions from a trajectory. 
The preference dataset $\mathcal{D}_{\rm p}=\{(\sigma_i^0, \sigma_i^1, y_i)\}_{i=1}^{|\mathcal{D}_{\rm p}|}$ consists of segments pairs with preference label given by the human annotators.  
The preference label is given by: $y_i=1$ if $\sigma^0 \succ \sigma^1$ and $y_i=0$ if $\sigma^1 \succ \sigma^0$, 
where we use  $\sigma \succ \sigma'$ to denote $\sigma$ is more preferred than $\sigma'$.  
We assume that all segments have the same length $|\sigma| = H$. 
The unlabeled dataset $\mathcal{D}_{\rm u}$, contains reward-free trajectories $\{\sigma_i\}_{i=1}^{|\mathcal{D}_{\rm u}|}$ collected by some behavior policy $\pi_\beta$. In practice, we usually have $|\mathcal{D}_{\rm u}| \gg |\mathcal{D}_{\rm p}|$ as collecting human annotations is more time-consuming and expensive compared to collecting unlabeled trajectories.

To learn a reward function $r$, a common approach is to assume a probabilistic  preference model and maximize the likelihood of the preference dataset,
\begin{align}
\mathcal{L} (\psi) = 
- \sum_{(\sigma^0, \sigma^1, y) \in \mathcal{D}_{\rm p}} (1-y) \log P(\sigma^0 \succ \sigma^1 ; \psi) + y P(\sigma^1 \succ \sigma^0 ; \psi)\, ,
\label{eq:lh}
\end{align}
where $P(\sigma \succ \sigma' ; \psi)$ is the preference model parameterized by the parameters $\psi$.  
For the probabilistic preference model, most existing methods adopt the Markovian reward assumption~\citep{christiano1, oprl,qpa}: 
\begin{align}
\rho_{\rm MR}(\sigma ; \psi) = \sum_{(s,a) \in \sigma} r_\psi (s, a)\, .
\label{eq:rho-mr}
\end{align}
That is, the \emph{preference strength} of a segment $\sigma$ correlates with its {cumulative rewards}. 
Applying the Bradley-Terry model~\citep{btmodel} leads to the following \emph{Markovian Reward (MR)} preference model,
\begin{align}
P_{\rm MR}(\sigma^0 \succ \sigma^1; \psi) =
\frac{\exp(\rho_{\rm {MR}} (\sigma^0 ; \psi ) )}{ \exp(\rho_{\rm {MR}} (\sigma^0  ; \psi )  ) + \exp(\rho_{\rm {MR} } (\sigma^1 ; \psi ) )} \, .
\label{eq:bt_model}
\end{align}
Plugging \eqref{eq:bt_model} into \eqref{eq:lh} yields a practical learning objective for learning the reward function.   
Finally, we can label $\mathcal{D}_{\rm u}$ with the learned reward model for any $(s,a)\in\sigma, \sigma\in\mathcal{D}_{\rm u}$. 
The resulting labeled dataset can be used for policy optimization with any offline RL algorithms, such as IQL  \citep{kostrikov2022offline} and AWAC \citep{nair2020awac}.

%% file: tex/4.method.tex
In this section we introduce a new preference model designed to address the limitations of the MR preference model by utilizing hindsight information.  
We begin with an illustrative example that serves as the primary motivation for our approach, followed by a detailed explanation of the formalization and implementation of the proposed method.


\subsection{Motivating Example: The Influence of the Future}
\label{sec:future_influence}

\begin{wrapfigure}{r}{0.5\textwidth}
  \begin{center}
    \includegraphics[width=0.85\linewidth]{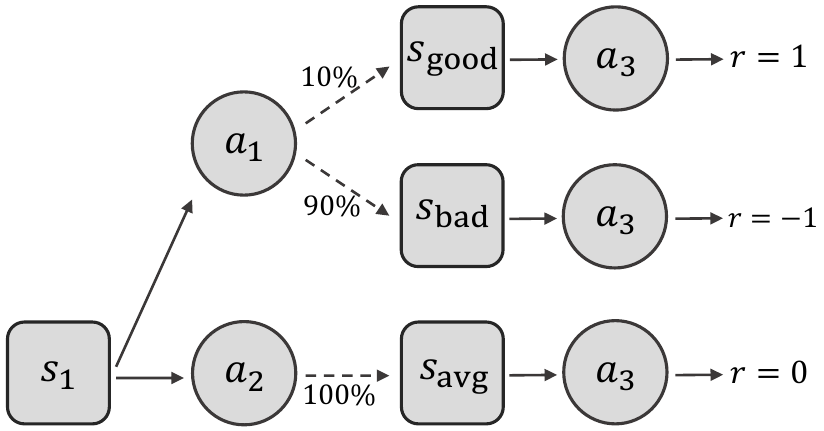}
  \end{center}
  \caption{A gambling MDP that illustrates the potential failure modes of the MR preference model. }
  \label{fig:gambling_mdp}
\end{wrapfigure}

To elucidate the influence of the future in preference modeling, we take the gambling MDP from~\citet{doc} as an example (Figure~\ref{fig:gambling_mdp}). 
An agent begins at $s_1$ with two actions: $a_1$ and $a_2$. Choosing the high-risk action $a_1$, the agent transitions to a rewarding state $s_{\rm good}$ with probability of $10\%$, but is more likely ($90\%$) to a penalizing state $s_{\rm bad}$. 
Alternatively, the safer and actually optimal action $a_2$ consistently leads to a neutral state $s_{\rm avg}$, yielding a reward of $0$.  
Suppose we are to extract the reward function using the provided dataset, where preferences are labeled based on the ground-truth reward:
$$
\begin{aligned}
    \mathcal{D}=\left\{
    \begin{aligned}
        &((s_1\rightarrow a_1 \rightarrow s_{\rm good}\rightarrow a_3), &(s_1\rightarrow a_2 \rightarrow s_{\rm avg}\rightarrow a_3),\quad &y=0) \\
        &((s_1\rightarrow a_1 \rightarrow s_{\rm good}\rightarrow a_3), &(s_1\rightarrow a_2 \rightarrow s_{\rm avg}\rightarrow a_3),\quad &y=0) \\
        &((s_1\rightarrow a_1 \rightarrow s_{\rm good}\rightarrow a_3), &(s_1\rightarrow a_1 \rightarrow s_{\rm bad}\rightarrow a_3),\quad &y=0) \\
        &((s_1\rightarrow a_1 \rightarrow s_{\rm bad}\rightarrow a_3), &(s_1\rightarrow a_2 \rightarrow s_{\rm avg}\rightarrow a_3),\quad &y=1) \\
    \end{aligned}\right\}.
\end{aligned}
$$
Applying the MR preference model \eqref{eq:bt_model} to this dataset would likely yield a reward function where  $r_\psi(s_1, a_1) > r_\psi(s_1, a_2)$,
because a larger proportion of trajectories involving $a_1$ lead to the rewarding state $s_{\rm good}$ (our experiments also validate this in Figure~\ref{fig:exp_gambling}). However, this violates rationality as selecting $a_2$ offers a higher return in expectation.
This failure can be attributed to the inappropriate credit assignment inherent in the MR preference model \eqref{eq:bt_model}:  
 a preference for $(s_1\rightarrow a_1\rightarrow s_{\rm good}\rightarrow a_3)$ will assign credits to both $r_\psi(s_1, a_1)$ and $r_\psi(s_{\rm good}, a_3)$ equally, leading to over-estimated utilities for $a_1$. To address this issue, a natural approach is to condition the reward of $(s_1, a_1)$ on the future outcome of the segment (i.e. $s_{\rm good}$ or $s_{\rm bad}$), so that $a_1$ is encouraged only when it leads to a favorable outcome $s_{\rm good}$. The conditional reward function $r_\psi(s, a | \sigma_{\rm future})$ thus answers the critical question: \textit{If my future is determined to be $\sigma_{\rm future}$, how advantageous it is for me to choose action $a$ at $s$? }



When applying $r_\psi(s, a | \sigma_{\rm future})$ to label data, we can marginalize over all possible future segments according to some prior distribution $p_{\rm prior}(\cdot |s, a) $ to get the value $r_\psi(s, a)=\mathbb{E}_{\sigma\sim p_{\rm prior}(\cdot|s, a)}[r_\psi(s, a| \sigma)]$. 
Note that the prior distribution can be estimated using the unlabeled offline dataset, which is unbiased concerning the environment transition and the behavior policy. Take the gambling MDP again as an example, the marginalization would decrease the utility of action $a_1$ since most of the time ($90\%$) the agent will arrive at the bad state.  
Besides, when there exists a mismatch between the data distribution of the preference dataset $\mathcal{D}_{\rm p}$ and that of the unlabeled dataset $\mathcal{D}_{\rm u}$ (which is a common occurrence because of the improvement of the policy or the non-uniform sampling for preference pairs), it is found that the reward from existing methods may not align with the RL agent's interests~\citep{qpa,rso}. 
The conditional reward model provides an effective solution to this issue by enabling the marginalization of the reward function $r_\psi$ using a data distribution that is more closely aligned with the on-policy data, thereby improving the alignment between learned behaviors and desired outcomes. 

\subsection{Hindsight Preference Model}\label{sec:hpm}

We now present \emph{Hindsight Preference Model (HPM)}, a novel preference model that incorporates future information in preference modeling. 
As opposed to \eqref{eq:rho-mr}, HPM assumes that the preference strength of a trajectory segment $\sigma=(s_0, a_0, s_1, a_1,\dots, s_H)$ is determined by 
\begin{align}
\rho_{\rm HPM}(\sigma ; \psi) = \sum_{(s_t, a_t) \in \sigma} r_\psi (s_t, a_t | \sigma_{t:t+k})\, ,
\label{eq:rho-hpm}
\end{align}
where $\sigma_{i:j}$ denote the subsequence of $\sigma$ between step $i$ and $j$.\footnote{We use $\sigma_{t:t+k}$ to simplify the notations. For $t+k > H$, the subsequence will be clipped to $\sigma_{t:\min(t+k, H)}$. } 
That is, in HPM the reward function $r_\psi$ not only takes the current state $s_t$ and action $a_t$ as input, but also depends on the $k$-step future outcome $\sigma_{t:t+k}$. 
Then given a  segment pair $(\sigma^0, \sigma^1)$, HPM model their preference by
\begin{align}
P_{\rm HPM}(\sigma^0 \succ \sigma^1; \psi) =
\frac{\exp(\rho_{\rm {HPM}} (\sigma^0 ; \psi ) )}{ \exp(\rho_{\rm {HPM}} (\sigma^0  ; \psi )  ) + \exp(\rho_{\rm {HPM} } (\sigma^1 ; \psi ) )} \, .
\label{eq:hpm}
\end{align}
In practice, directly implementing this conditional reward $r_\psi (s_t, a_t | \sigma_{t:t+k})$ is challenging due to the high-dimensional nature of the $k$-step segment $\sigma_{t:t+k}$. 
We address this issue by compressing the segment into a compact embedding rather than operating directly in the raw space of trajectory.

\begin{algorithm}[t]
\caption{Hindsight Preference Learning (HPL)}
\label{algo:hpl}
\textbf{Input}: Preference dataset $\mathcal{D}_{\rm p}=\{(\sigma_i^0, \sigma_i^1, y_i)\}_{i=1}^{|\mathcal{D}_{\rm p}|}$, unlabeled dataset $\mathcal{D}_{u}=\{\sigma_i\}_{i=1}^{|\mathcal{D}_{\rm u}|}$.

\begin{algorithmic}[1]
\STATE {\color{gray}// VAE Training}
\STATE Initialize the segment encoder $q_\theta$, decoder $p_\theta$ and prior $f_\theta$ with parameters $\theta$.
\FOR{$n=1, 2, \cdots, N_{\rm VAE}$}
\STATE Sample minibatch of segments $\sigma\sim \mathcal{D}_{u}$
\STATE Update $\theta$ by maximizing Eq~\eqref{eq:elbo}
\ENDFOR
\STATE {\color{gray}// Reward Learning}
\STATE Initialize the reward function $r_\psi$ with parameters $\psi$
\FOR{$n=1, 2, \cdots, N_{\rm HPM}$}
\STATE Sample minibatch of preference pairs $(\sigma^0, \sigma^1, y)\sim \mathcal{D}_{p}$
\STATE Update $\psi$ by minimizing Eq~\eqref{eq:lh} with $P_{\rm HPM}$ defined in Eq~\eqref{eq:empirical_hpm}
\ENDFOR
\STATE {\color{gray}// RL Training}
\STATE Label the reward for $\mathcal{D}_{\rm u}$ using Eq~\eqref{eq:marginalization} and optimize the policy with any offline RL algorithm
\end{algorithmic}
\end{algorithm}

\subsection{Future Segment Embedding}\label{sec:traj_embedding}

We propose to compress future segments $\sigma_{t:t+k}$ into a compact embedding vector $z_t$ by training a conditional \textit{Variational Auto-Encoder} (VAE) \citep{kingma2013auto}. 
The architecture of our model consists of three components: the encoder $q_\theta$, the decoder $p_\theta$, and a learnable prior $f_\theta$, which can be jointly optimized with the \emph{Evidence Lower Bound (ELBO)}:
\begin{equation}\label{eq:elbo}
    \begin{aligned}
        &\log p(\sigma_{t:t+k}|s_t, a_t)\\
        &\geq \mathbb{E}_{q_\theta(z_t|s_t, a_t, \sigma_{t:t+k})}\left[\log \frac{p_\theta(\sigma_{t:t+k}, z_t|s_t, a_t)}{q_\theta(z_t|s_t, a_t, \sigma_{t:t+k})}\right]\\
        &=\mathbb{E}_{q_\theta(z_t|s_t, a_t, \sigma_{t:t+k})}\left[\log p_\theta(\sigma_{t:t+k}|s_t, a_t, z_t)\right] - \operatorname{KL}\left[q_\theta(z_t|s_t, a_t, \sigma_{t:t+k})\|f_\theta(z_t|s_t, a_t)\right] \\
        &\stackrel{\text{def}}{=} - \mathcal{L}_{\rm ELBO}(s_t, a_t, \sigma_{t:t+k}; \theta).
    \end{aligned}
\end{equation}


Following the pre-training phase, the VAE can be utilized for both reward learning and reward labeling. 
In the context of reward learning, the embedding $z_t$ can be employed as a substitute for $\sigma_{t:t+k}$ in preference modeling:
\begin{align}
\rho_{\rm HPM}(\sigma ; \psi) = \sum_{(s_t, a_t) \in \sigma} r_\psi (s_t, a_t | \sigma_{t:t+k})
\approx
\sum_{(s_t, a_t) \in \sigma} r_\psi (s_t, a_t | z_t)
\, .
\label{eq:empirical_hpm}
\end{align}
Here, the embedding is obtained using the encoder $z_t \sim q_\theta(\cdot | s_t, a_t, \sigma_{t:t+k})$. 
Plugging this into the Bradley-Terry model gives an approximation of HPM \eqref{eq:hpm}. 
We then once again utilize the preference dataset along with the cross-entropy loss, as defined in \eqref{eq:lh}, to optimize $r_\psi$.  
During the reward labeling phase, 
we compute the reward using the prior distribution $f_\theta$:
\begin{equation}\label{eq:marginalization}
\begin{aligned}
    r_\psi(s_t, a_t) = \mathbb{E}_{z_t\sim f_\theta(\cdot|s_t, a_t)}\left[r_\psi(s_t, a_t, z_t)\right] 
    \approx \frac{1}{N} \sum_{l=1}^N r_\psi(s_t, a_t, z_t^{(l)}) \, ,
\end{aligned}
\end{equation}
where $z_t^{(1)}, z_t^{(2)}, \ldots, z_t^{(N)}$ are i.i.d. samples from $f_\theta$. This approach ensures a robust approximation of the expected reward, facilitating effective reward shaping based on learned preferences.

We train these models using the unlabelled dataset $\mathcal{D}_{\rm u}$. This offers two benefits. Firstly, $\mathcal{D}_{\rm u}$ typically encompasses a substantial volume of data, which enhances model performance. Secondly, the scalar reward is obtained by marginalizing over the prior distribution $f_\theta$ during the reward labeling phase. Precisely aligning $f_\theta$ with  $\mathcal{D}_{\rm u}$ can significantly enhance the stability of this marginalization process, particularly in instances of distributional shifts between $\mathcal{D}_{\rm u}$ and $\mathcal{D}_{\rm p}$.  

\paragraph{Practical Implementation.}

We employ the GPT architecture \citep{brown2020language} for the encoder $q_\theta$ due to its expressivity in sequence modeling.  
Given an input segment 
$\sigma=(s_0, a_0, s_1, a_1,\dots, s_{H}, a_{H})$, 
we concatenate $s_t$ and $a_t$ together as a single token. In each attention layer, we apply the anti-causal attention mask which restricts each token's attention to itself and subsequent tokens, ensuring that the output token $z_t$ encapsulates the forward-looking information starting from time step $t$. 
The decoder network $p_\theta$ reconstructs $\sigma_{t:t+k}$ using the embedding $z_t$ and $(s_t, a_t)$. 
In our implementation, $p_\theta$ takes inputs of $s_t, a_t, z_t$ and a time interval $\Delta t\in[0, 1, ..., k]$, and predicts $(s_{t+\Delta t}, a_{t+\Delta t})$. This parameterization facilitates the parallel decoding of the entire trajectory by processing all specified intervals in a single forward pass. Finally, the prior $f_\theta$ is parameterized as an MLP network which receives $(s_t, a_t)$ and outputs a distribution over the embedding space. 

\subsection{Overall Framework of HPL}\label{sec:overall}

Putting everything together, we outline \emph{Hindsight Preference Learning (HPL)} in Algorithm~\ref{algo:hpl}. HPL can be divided into three stages: 1) pre-training a VAE to embed future segments, using data from the unlabeled dataset $\mathcal{D}_{\rm u}$; 2) training the conditional reward function $r_\psi$ with the preference dataset $\mathcal{D}_{\rm p}$; and finally 3) label the unlabeled dataset with \eqref{eq:marginalization}, followed by applying any offline RL algorithms for policy optimization.

%% file: tex/2.related_work.tex
\textbf{Preference-based Reinforcement Learning. }Human preferences are easier to obtain compared to well-calibrated step-wise rewards or expert demonstrations in some domains, making them rich yet easy source of signals for policy optimization. \citet{christiano1} utilizes the Bradley-Terry model to extract reward function from human preferences and lays the foundation of using deep RL to solve complex tasks. Based on this, several methods~\citep{pebble,pbrl1,pbrl2} further improves the query efficiency by incorporating techniques like pre-training and relabeling. OPRL~\citep{oprl} further proposes principled rules for query selection and provides baseline results using existing offline datasets. On the other hand, some works bypass the need for a reward model. IPL~\citep{ipl} achieves this by expressing the reward with value functions via the inverse Bellman operator, while OPPO~\citep{oppo} uses hindsight information matching (HIM) to conduct preference learning in compact latent space. FTB~\citep{ftb} employs powerful generative models to diffuse bad trajectories to preferred ones. DPPO~\citep{dppo} and CPL~\citep{cpl}, although with different starting points, both directly optimize the policy by relating it to the preferences.

\textbf{Human Preference Modeling. }To extract utilities from human preferences for policy optimization, we need preference models to establish the connection between preferences and utilities. A common approach is to use the Bradley-Terry model~\citep{christiano1} and hypothesizes that preference is emitted according to the sum of Markovian rewards, while Preference Transformer~\citep{pt} and Hindsight PRIOR~\citep{prior} extend this by using the weighted sum of non-Markovian rewards. Besides, another line of research proposes that human preference is decided by the sum of optimal advantages in the segment~\citep{regret_rm, learning_adv}, rather than the rewards. In this paper, we focus on the influence of the future and consider the sum of future-conditioned rewards. 

\textbf{Leveraging Hindsight Information. }Hindsight information can provide extra supervision during training. For example, HER~\citep{her} proposes to relabel the transitions to allow sample-efficient learning in sparse-reward tasks. Prior works have also explored learning representations by predicting the future~\citep{generalized_dt, pdt, doc}, and such representations facilitate downstream tasks such as policy optimization~\citep{generalized_dt, pdt}, preference modeling~\citep{oppo}, exploration~\citep{curiosity_in_hindsight}, and credit assignment~\citep{hca}.

%% file: tex/5.experiments.tex
We evaluate HPL as well as other methods with various benchmarks. Specifically, we selected two tasks (Hopper and Walker2D) from Gym-MuJoCo locomotion~\citep{brockman2016openai}, two tasks (Hammer and Pen) from the Adroit manipulation platform~\citep{Kumar2016thesis}, and four tasks (Drawer-Open, Button-Press, Plate-Slide and Sweep-Into) from Meta-World Benchmark~\citep{yu2020meta}. For Gym-MuJoCo tasks and Adroit tasks, we select datasets from the D4RL Benchmark~\citep{d4rl} and mask the reward labels as the unlabeled dataset $\mathcal{D}_{\rm u}$, while the annotated preference dataset $\mathcal{D}_{\rm p}$ is provided by \citet{pt}. For Meta-World tasks, we used the datasets released by \citet{ipl} as $\mathcal{D}_{\rm u}$ and $\mathcal{D}_{\rm p}$. It is worthwhile to note that for Gym-MuJoCo and Adroit tasks, the preference label is generated by real human annotators, while for Meta-World tasks it is synthesized based on trajectory return. More details about the datasets and how the preference annotations are generated can be found in Appendix~\ref{appsec:dataset}. 

For baseline methods, we consider 1) \textbf{Oracle}, which uses the oracle step-wise reward for policy optimization; 2) Supervised Fine-Tuning (\textbf{SFT}), which imitates the preferred segments; 3) \textbf{MR}, which uses the Bradley-Terry Model to extract Markovian rewards from the preferences; 4) Preference Transformer (\textbf{PT})~\citep{pt}, which uses a transformer and bidirectional layer to model the reward; and 5) Inverse Preference Learning (\textbf{IPL})~\citep{ipl}, which removes the need of reward modeling using inverse Bellman operator. More details about the implementations can be found in Appendix~\ref{appsec:implementation}. For evaluation metrics, we report the normalization score calculated by D4RL and the success rate for Meta-World tasks. 

\input{tables/main_result.tex}

\vspace{-2mm}
\subsection{Benchmark Results}\label{sec:main_results}
Our first experiment investigates the capability of HPL in standard offline PbRL settings using both $\mathcal{D}_{\rm u}$ and $\mathcal{D}_{\rm p}$. For policy optimization, we used IQL for all methods except for SFT. We found certain design choices such as reward normalization can have a significant effect on the performance, so we included the reference score (denoted as ref.) from the original paper for some algorithms and the score of our implementation (denoted as reimpl.) for fair comparisons.

The results are listed in Table~\ref{tab:main_result}. We also implement variants that use AWAC for policy optimization, and the results are deferred to Appendix~\ref{appsec:awac}. Overall, HPL consistently outperforms other baselines both in locomotion tasks and manipulation tasks, especially in complex domains like the pen task. The promising performance validates the effectiveness of HPL for learning from human preferences.

\subsection{Tasks with Preference Distribution Shift}
\begin{wrapfigure}{r}{0.6\textwidth}
  \begin{center}
    \includegraphics[width=1.0\linewidth]{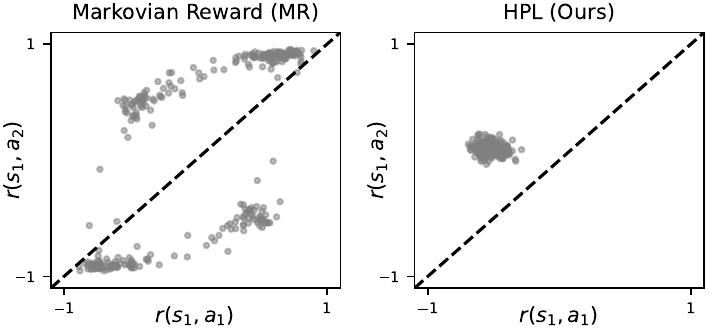}
  \end{center}
  \caption{The rewards values given by the MR method and HPL. Each dot represents one trial and its coordinates are the estimated reward values.}
  \label{fig:exp_gambling}
\end{wrapfigure}

As we illustrated in Section~\ref{sec:future_influence}, the distribution mismatch between the preference dataset $\mathcal{D}_{\rm p}$ and the unlabeled dataset $\mathcal{D}_{\rm u}$ may affect credit assignments. We take the gambling MDP (Figure~\ref{fig:gambling_mdp}) as a sanity check to see whether HPL can deliver better results. We used the dataset $\mathcal{D}$ in Section~\ref{sec:future_influence} as the preference dataset $\mathcal{D}_{\rm p}$, and additionally collected $\mathcal{D}_{u}$ by randomly choosing between $a_1$ and $a_2$. Afterwards, we compare the reward values of $(s_1, a_1)$ and $(s_1, a_2)$ given by both MR and HPL. We run both algorithms for 500 random seeds and plot the results in Figure~\ref{fig:exp_gambling}. In the figure, each point stands for one trial and its coordinate stands for $r_\psi(s_1, a_1)$ and $r_\psi(s_1, a_2)$ respectively. We note that every trial of both methods has achieved $100\%$ accuracy for predicting the preference labels, so we focus on the quality of rewards. As discussed in Section~\ref{sec:future_influence}, successful credit assignment should try to assign lower values to $r(s_1, a_1)$, i.e. the point should lie in the above triangular area. While HPL steadily assigns higher rewards to $(s_1, a_2)$, MR over-estimates the rewards for $(s_1, a_1)$ in almost half of the cases. Although it is still possible that MR can find the optimal policy through RL optimization, we emphasize that HPL delivers a much more robust and advantageous utility value that will facilitate the subsequent policy optimization. 
\vspace{-3mm}

\begin{figure*}[htbp]
    \centering
    \includegraphics[width=0.88\linewidth]{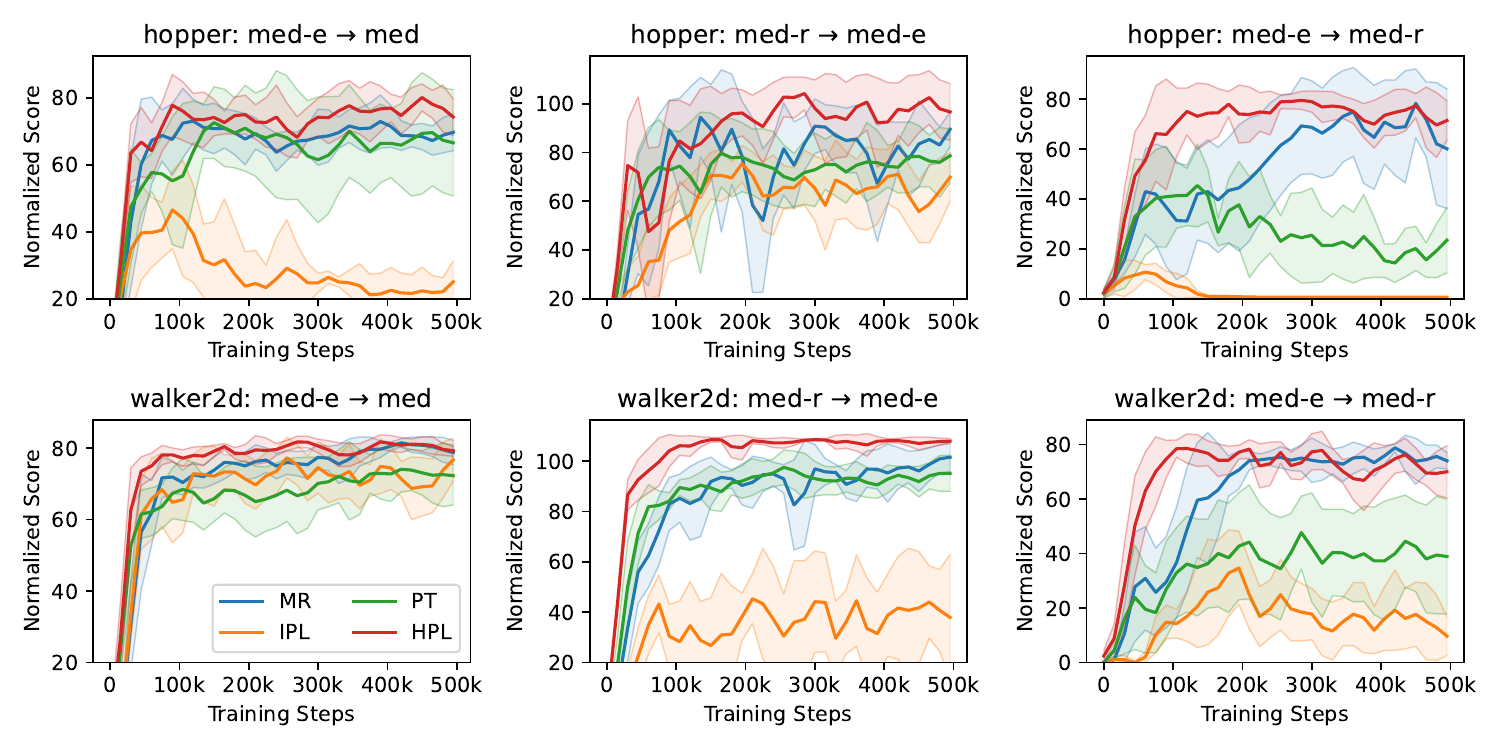}
    \caption{The performance curves of HPL and baseline methods in tasks with mismatched datasets. 
    We report the average (solid line) and the standard deviation (shaded area) of each algorithm across 5 random seeds and 10 evaluation episodes for each seed.}
    \label{fig:exp_mismatch}
\end{figure*}

To validate HPL's capability of learning better rewards in the face of the preference distribution shift on a larger scale, we constructed a series of tasks by combining $\mathcal{D}_{\rm u}$ and $\mathcal{D}_{\rm p}$ collected by different behavior policies. For example, "hopper: med-e $\rightarrow$ med" means we used $\mathcal{D}_{\rm p}$ with \textit{medium-expert} quality and $\mathcal{D}_{\rm u}$ with \textit{medium} quality. The performance curves are presented in Figure~\ref{fig:exp_mismatch}. In such mismatched scenarios, HPL performs better than all baseline algorithms in most of the tasks. Besides, HPL demonstrates faster convergence speeds and more stable performances, which validates that the reward from HPL is more robust and shaped for downstream RL optimization. 

\vspace{-2mm}
\subsection{Analysis of HPL}
In this section, we examine each part of HPL to gain a deeper understanding of each design choice. 

\begin{figure*}[htbp]
    \centering
    \begin{minipage}{0.55\linewidth}
        \centering
        \includegraphics[width=1.0\linewidth]{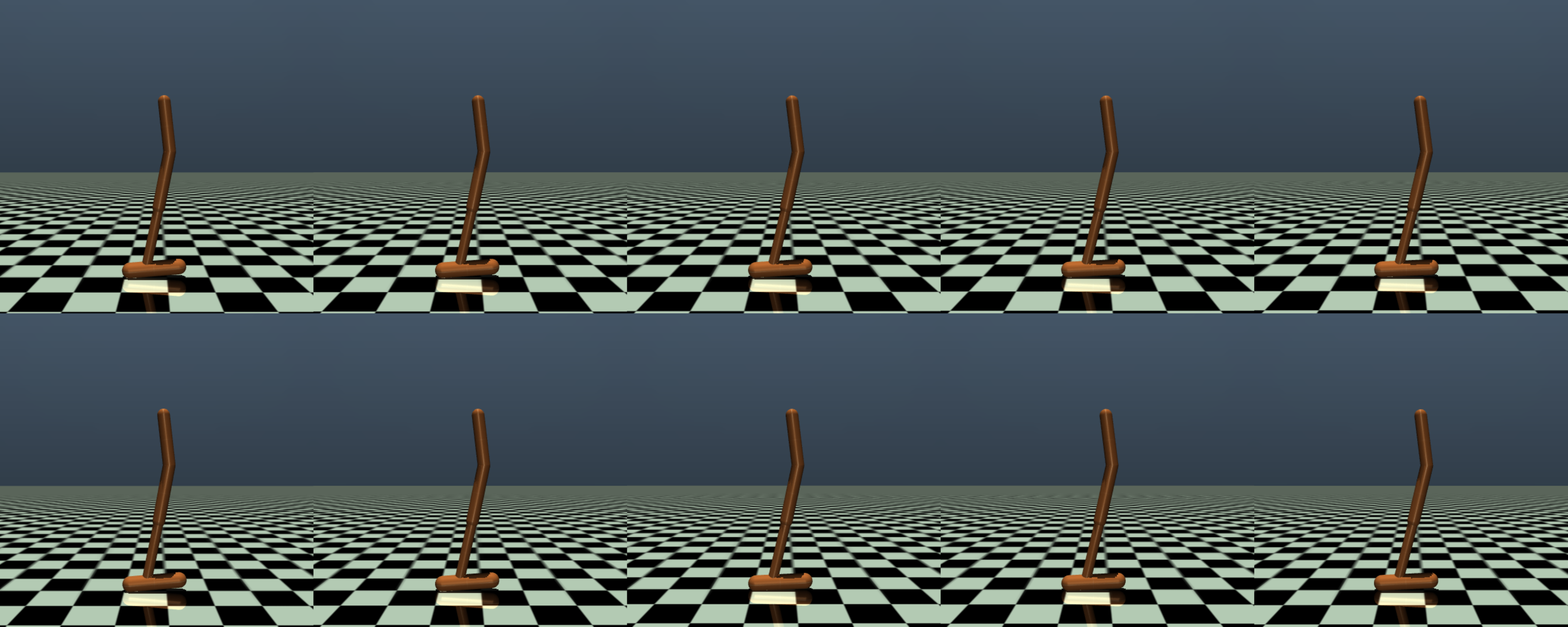}
    \end{minipage}%
    \begin{minipage}{0.3\linewidth}
        \centering
        \includegraphics[width=1.0\linewidth]{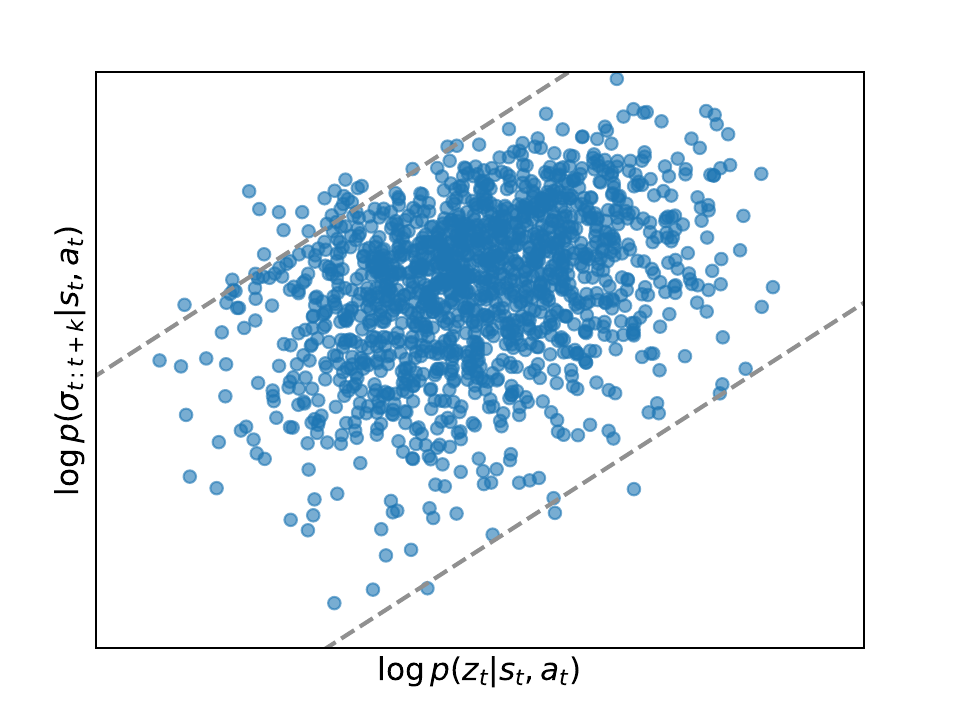}
    \end{minipage}
    \caption{Left: The rendered image of one raw trajectory selected from the offline dataset (top row) and the reconstruction by the VAE (bottom row). Right: The relationship between the log-probabilities of segments and their embeddings. }\label{fig:vae}
\end{figure*}

\textbf{Future Segment Embedding and the VAE Structure. }HPL relies on the VAE structure to generate compact embeddings for future segment representation and sampling. Consequently, our first analysis investigates the quality of these embeddings. 
Figure~\ref{fig:vae} displays the images of a trajectory segment from the offline dataset (top left) and its reconstruction by the VAE (bottom left). It is important to note that both the encoding and reconstruction processes are based on states and actions, rather than pixel observations. The VAE reconstruction is highly accurate, indicating that the embedding $z_t$ effectively compresses the relevant information. 
In the right figure, we select one $(s_t, a_t)$ from the offline dataset and simulate trajectories from that point using a behavior cloning policy $\pi_\beta$. Subsequently, we plot the relationship between the log probability of each trajectory $\log p(\sigma_{t:t+k}|s_t, a_t)=\sum_{i=1}^k \log \pi_\beta(a_{t+i}|s_{t+i})$) and the log probability of the embedding $\log p(z_t|s_t, a_t)=\log f_\theta(z_t|s_t, a_t)$). The positive correlation observed between these two log validates the efficacy of sampling from $f_\theta$.

\begin{figure*}[htbp]
    \centering
    \begin{subfigure}[t]{0.28\textwidth}
        \centering
        \includegraphics[width=0.98\linewidth]{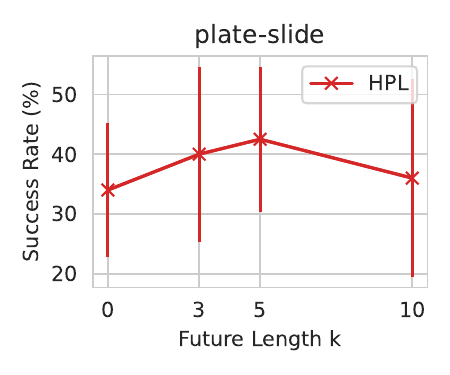}
        \caption{Ablation on $k$}
        \label{fig:exp_future_len}
    \end{subfigure}%
    \begin{subfigure}[t]{0.28\textwidth}
        \centering
        \includegraphics[width=1.0\linewidth]{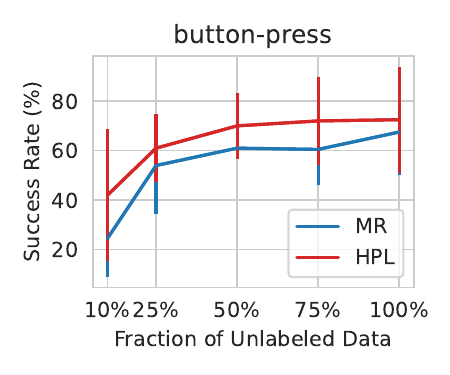}
        \caption{Scaling trend for $\mathcal{D}_{\rm u}$}
        \label{fig:exp_uscale}
    \end{subfigure}
    \begin{subfigure}[t]{0.28\textwidth}
        \centering
        \includegraphics[width=1.0\linewidth]{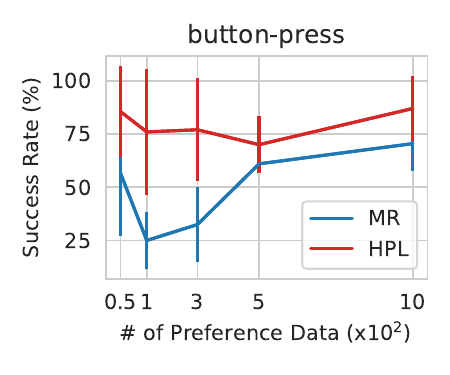}
        \caption{Scaling trend for $\mathcal{D}_{\rm p}$}
        \label{fig:exp_pscale}
    \end{subfigure}%
    \caption{Quantitative analysis of HPL. We report the mean (solid line) and the standard deviation (error bar) across 10 seeds for all experiments. }
\end{figure*}

\textbf{Ablation on the Future Length $k$. }The parameter $k$ controls the length of future segments encoded into the embedding $z$. As illustrated in Figure~\ref{fig:exp_future_len} (full tasks in Figure~\ref{fig:app_future_len}), extending $k$ generally enhances performance, highlighting the benefits of future conditioning. However, as $k$ exceeds a certain threshold, we witness \revise{increased fluctuation of the performance}{fluctuations and decreases in the performances}, probably due to the challenges of representing longer future segments accurately.

\vspace{-2mm}
\vspace{-2mm}

\textbf{Scaling with Dataset Sizes. }We conduct experiments to evaluate the scalability of MR and HPL with varying dataset sizes. In Figure~\ref{fig:exp_uscale} (full tasks in Figure~\ref{fig:app_uscale}), we adjust the size of $\mathcal{D}_{\rm u}$ from 10\% to 100\% of its total capacity and observe that HPL consistently outperforms MR across all proportions of unlabeled data. Figure~\ref{fig:exp_pscale} (full tasks in Figure~\ref{fig:app_pscale}) illustrates the scaling trends of HPL and MR with different numbers of preference queries. While both methods exhibit upwards trend in success rates, HPL \revise{demonstrates a more significant improvement}{is significantly better}. These experiments collectively confirm the scalability of HPL.

%% file: tables/main_result.tex
\begin{table*}[htbp]
\caption{Normalized averaged score for locomotion tasks (top) and manipulation tasks (bottom). In the table, "hop" is abbreviated for the Hopper task, "walk" for Walker2D, "ham" for Hammer, "m" for medium, "r" for replay, "e" for expert, "h" for human, "c" for cloned. The reference scores for MR and PT are from \citet{pt}, while those for IPL are from \citet{ipl}. For the rest numbers, we use our own implementations and report the average and the standard deviation of the performances across 10 evaluation episodes and 5 seeds. We bolded values that are within $95\%$ of the top-performing method among our implemented versions.}
\label{tab:main_result}
\centering
\setlength{\tabcolsep}{1.2mm}{}
\begin{small}
\begin{tabular}{cccccccccc}
\toprule[1.5pt]
\multicolumn{1}{c}{\multirow{2}{*}{Dataset}} & \multicolumn{1}{c}{\multirow{2}{*}{\textbf{Oracle}}}& \multicolumn{1}{c}{\multirow{2}{*}{\textbf{SFT}}}&\multicolumn{2}{c}{\textbf{MR}} & \multicolumn{2}{c}{\textbf{PT}} & \multicolumn{2}{c}{\textbf{IPL}} & \multicolumn{1}{c}{\multirow{2}{*}{\textbf{HPL}}} \\
&&&\multicolumn{1}{c}{ref.} & \multicolumn{1}{c}{reimpl.} & \multicolumn{1}{c}{ref.} & \multicolumn{1}{c}{reimpl.} & \multicolumn{1}{c}{ref.} & \multicolumn{1}{c}{reimpl.} &\\
\midrule
hop-m-r  & $97.4$ &$22.2$& $11.56$ & $64.3_{\pm 18.2}$ & $84.54$ & $77.4_{\pm 8.0}$ & $73.57$ & $56.1_{\pm 20.3}$ & \revise{$\bm{85.0_{\pm 5.3}}$}{$\bm{83.0_{\pm 14.4}}$} \\
hop-m-e  & $107.4$ &$5.2$& $57.75$ & $86.3_{\pm 21.6}$ & $68.96$ & $78.7_{\pm 27.8}$ & $74.52$ & $67.8_{\pm 18.0}$ & \revise{$\bm{109.9_{\pm 2.6}}$}{$\bm{104.0_{\pm 7.7}}$} \\
walk-m-r & $82.2$ &$9.0$& $72.07$ & $\bm{69.5_{\pm 1.7}}$ & $71.27$ & $64.0_{\pm 15.2}$ & $59.92$ & $42.3_{\pm 17.4}$ & \revise{$57.7_{\pm 15.4}$}{$64.1_{\pm 8.9}$} \\
walk-m-e & $111.7$&$0.4$& $108.32$& $90.8_{\pm 9.0}$ & $110.13$& \revise{$\bm{102.2_{\pm 17.5}}$}{$102.2_{\pm 17.5}$} & $108.51$ & $\bm{106.1_{\pm 4.6}}$ & \revise{$\bm{105.8_{\pm 8.0}}$}{$\bm{108.9_\pm{0.5}}$}\\
\midrule\midrule

\multicolumn{1}{c}{Dataset} & \multicolumn{1}{c}{\textbf{Oracle}}& \multicolumn{1}{c}{\textbf{SFT}}&\multicolumn{2}{c}{\textbf{MR}} & \multicolumn{2}{c}{\textbf{PT}} & \multicolumn{2}{c}{\textbf{IPL}} & \multicolumn{1}{c}{\textbf{HPL}} \\
\midrule
pen-h    & $78.5$ &$36.4$&\multicolumn{2}{c}{$14.1_{\pm 9.0}$}&\multicolumn{2}{c}{$11.2_{\pm 4.5}$}&\multicolumn{2}{c}{$11.5_{\pm 11.6}$}&\revise{$\bm{62.0_{\pm 24.2}}$}{$\bm{70.9_{\pm 23.2}}$} \\
pen-c    & $83.4$ &$31.1$&\multicolumn{2}{c}{$13.8_{\pm 4.8}$}&\multicolumn{2}{c}{$11.9_{\pm 13.3}$}&\multicolumn{2}{c}{$12.3_{\pm 6.6}$}&\revise{$\bm{32.3_{\pm 17.0}}$}{$\bm{33.1_{\pm 19.6}}$} \\
ham-h    & $1.8$  &$0.3$&\multicolumn{2}{c}{$0.2_{\pm 0.0}$}&\multicolumn{2}{c}{$0.2_{\pm 0.3}$}&\multicolumn{2}{c}{$0.0_{\pm 0.0}$}&\revise{$\bm{3.4_{\pm 2.8}}$}{$\bm{4.3_{\pm 4.7}}$} \\
ham-c    & $1.5$  &$\bm{2.6}$&\multicolumn{2}{c}{$0.0_{\pm 0.1}$}&\multicolumn{2}{c}{$2.0_{\pm 4.6}$}&\multicolumn{2}{c}{$0.1_{\pm 0.1}$}&\revise{$0.2_{\pm 0.0}$}{$0.3_{\pm 0.0}$} \\
drawer-open  & - &$0.42$&\multicolumn{2}{c}{$\bm{0.92_{\pm 0.10}}$}&\multicolumn{2}{c}{$0.39_{\pm 0.24}$}&\multicolumn{2}{c}{$0.54_{\pm 0.26}$}&\multicolumn{1}{c}{\revise{$\bm{0.92_{\pm 0.13}}$}{$\bm{0.95_{\pm 0.07}}$}} \\
button-press & - &$0.26$&\multicolumn{2}{c}{$0.61_{\pm 0.04}$}&\multicolumn{2}{c}{$0.38_{\pm 0.24}$}&\multicolumn{2}{c}{$0.58_{\pm 0.11}$}&\multicolumn{1}{c}{\revise{$\bm{0.84_{\pm 0.17}}$}{$\bm{0.70_{\pm 0.14}}$}} \\
plate-slide  & - &$0.26$&\multicolumn{2}{c}{\revise{$\bm{0.38_{\pm 0.08}}$}{$0.38_{\pm 0.08}$}}&\multicolumn{2}{c}{$0.29_{\pm 0.27}$}&\multicolumn{2}{c}{$0.34_{\pm 0.29}$}&\multicolumn{1}{c}{\revise{$0.28_{\pm 0.04}$}{$\bm{0.43_{\pm 0.13}}$}} \\
sweep-into   & - &$0.24$&\multicolumn{2}{c}{$0.31_{\pm 0.10}$}&\multicolumn{2}{c}{$0.22_{\pm 0.13}$}&\multicolumn{2}{c}{$0.14_{\pm 0.15}$}&\multicolumn{1}{c}{\revise{$\bm{0.33_{\pm 0.10}}$}{$\bm{0.37_{\pm 0.11}}$}} \\

\bottomrule[1.5pt]
\end{tabular}
\end{small}
\end{table*}

%% file: tex/6.conclusions.tex
This paper focuses on extracting rewards from human preferences for RL optimization. Unlike previous methods that assume the preference is determined by the sum of Markovian rewards, our method, HPL, instead employs a new preference model that correlates the preference strength with the sum of rewards which are conditioned on the future outcome of this trajectory. By marginalizing the conditional reward over the prior distribution of future outcomes induced by the vast unlabeled dataset, HPL produces more robust and suitable reward signals for downstream RL optimization. 

\textbf{Limitations. }The primary limitation of the current version of HPL lies in its failure to exploit the full potential of the learned VAE. Although the VAE functions as a generative model, it has not been employed to augment the unlabeled offline dataset through sampling.  Additionally, the VAE architecture holds potential for other capabilities, such as uncertainty estimation~\citep{liang2022reward} and the identification of diverse preferences~\citep{xue2023reinforcement}, which have yet to be explored. Future work will focus on extending the HPL framework to fully harness these capabilities and thereby improve its overall performance.

%% file: tex/appendix.tex
\section{Tasks and Datasets}\label{appsec:dataset}

\subsection{Tasks}

We evaluated the HPL algorithm on different environments, including Gym-MuJoCo~\citep{brockman2016openai}, Adroit~\citep{Kumar2016thesis}, and Meta-World~\citep{yu2020meta}. These tasks range from basic locomotion to complex manipulation. Among them, Gym-MuJoCo and Meta-World are released with MIT license, while Adroit is released with the Apache-2.0 license. 

\textbf{Gym-MuJoCo.} We selected the Hopper and Walker2D tasks from the Gym-MuJoCo environment. The goal of the Hopper task is to control a single-legged robot to hop forward, with primary rewards based on forward speed, energy consumption, and a survival bonus for maintaining stability. The Walker2D task involves controlling a bipedal robot to walk forward, while the rewards are designed based on the forward speed, control penalties, and a survival bonus. The key challenge in both tasks is to maximize the forward distance while maintaining the robot's stability.

\textbf{Adroit.} We chose the Hammer and Pen tasks from the Adroit environment. These tasks require controlling a 24-DoF simulated Shadow Hand robot to perform precise manipulations. The Hammer task involves using the robot to hammer a nail, with rewards given for successful strikes and penalties for misses or ineffective actions. The Pen task requires the robot to rotate a pen, rewarding successful rotations and penalizing failures or instability. Adroit tasks emphasize high precision and the complexity of robotic hand manipulations.

\textbf{Meta-World.} We selected multiple manipulation tasks from the Meta-World environment, including drawer-open, button-press, plate-slide, and sweep-into. These tasks require a Sawyer robotic arm to perform multi-step operations. For example, the drawer-open task involves grasping and pulling open a drawer, the button-press task requires accurately pressing a designated button, the plate-slide task involves pushing a plate to a specified location, and the sweep-into task requires sweeping objects into a target area. The reward structure in these tasks is designed as a combination of sub-tasks, providing partial rewards for each sub-task completed and a total reward for successfully completing the entire task. Meta-World tasks highlight the shared structure between tasks and the sequential nature of complex operations.

\subsection{Unlabeled Offline Datasets}
\begin{figure*}[htbp]
    \centering
    \includegraphics[width=1.0\linewidth]{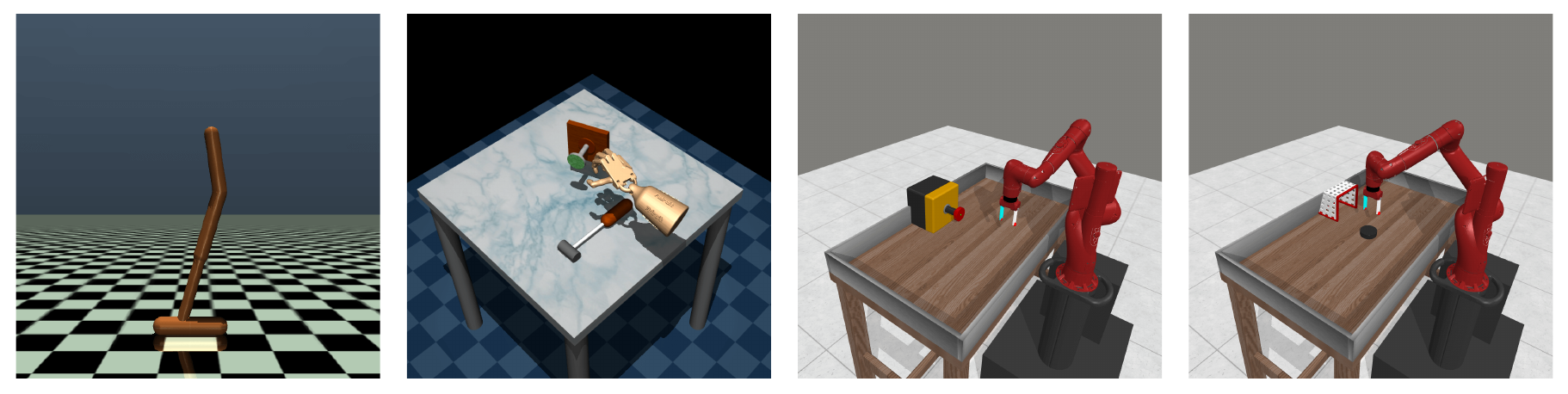}
    \caption{From left to right, the figures show the Hopper task from Gym-MuJoCo, the Hammer task from the Adroit platform, and the button-press and plate-slide tasks from Meta-World. }
\end{figure*}

For the unlabeled offline dataset $\mathcal{D}_{\rm p}$, we used the datasets provided in D4RL~\citep{d4rl} for Gym-MuJoCo and Adroit tasks and the datasets from \citet{ipl} for Meta-World tasks.

\textbf{Gym-MuJoCo Datasets.} The datasets for Hopper and Walker2D tasks were obtained through online training and include \emph{medium}, \emph{medium-replay}, and \emph{medium-expert} datasets. The \emph{medium} dataset was generated by training a policy using Soft Actor-Critic~\citep{sac}, stopping early when the policy reached a medium performance level, and collecting 1 million samples from this partially trained policy. The \emph{medium-replay} dataset contains all samples observed in the replay buffer during training until the policy reaches medium performance. The \emph{medium-expert} dataset was created by mixing equal amounts of expert demonstration data and medium-level policy data, aiming to test the algorithm's performance with varying data quality. All of these datasets can be obtained following the APIs provided by D4RL, and the datasets are licensed with the CC BY 4.0 license. 

\textbf{Adroit Datasets.} We used data for the Hammer and Pen tasks. These datasets include \emph{human}, \emph{expert}, and \emph{cloned} datasets. The \emph{human} dataset consists of a small number of demonstrations collected from human experts, with each task containing 25 trajectories. The \emph{expert} dataset comprises a large amount of expert data collected from fine-tuned RL policies. The \emph{cloned} dataset is generated by training an imitation policy on the demonstration data, running this policy, and mixing the generated data with the original demonstrations in a 50-50 ratio. This data generation method simulates a real-world scenario where a small amount of human demonstration data is augmented through imitation learning. All of these datasets can be obtained following the APIs provided by D4RL, and the datasets are licensed with the CC BY 4.0 license. 

\textbf{Meta-World Datasets.} We used the preference-annotated dataset from \cite{ipl} and converted it into an unlabeled offline dataset by discarding the reference label. These datasets can be obtained following the original source of \cite{ipl}. The authors did not specify the license of the datasets. However, the codes are released with the MIT license so we speculate the datasets inherit the MIT license as well since they are released together. In the following paragraphs, we detail the construction of these preference datasets based on the details provided by their original creators. 

\subsection{Preference Datasets}
For the preference dataset $\mathcal{D}_{\rm u}$, we selected the human-annotated datasets from \citet{pt} for Gym-MuJoCo and Adroit tasks, and the synthetic datasets from \citet{ipl} for Meta-World tasks. The datasets are released alongside with the codes (\url{https://github.com/csmile-1006/PreferenceTransformer} and \url{https://github.com/jhejna/inverse-preference-learning} respectively). The authors did not specify the license of the datasets. However, the codes are both released with the MIT license so we speculate the datasets inherit the MIT license as well since they are released together. In the following paragraphs, we detail the construction of these preference datasets based on the details provided by their original creators.

For Gym-MuJoCo datasets, preferences were collected from actual human subjects. Specifically, human annotators watched the rendered videos of segments and selected the segment they believed was more helpful in achieving the agent's goal. Each segment lasted 3 seconds (100 frames). Human annotators can prefer one of the segment pairs or remain neutral by assigning equal preference to both segments. 
The annotators are instructed to make decisions based on some criteria. For the Hopper task, the robot is expected to move to the right as far as possible while minimizing energy consumption. Segments, where the robot lands unstably, are rated lower, even if the distance traveled is longer. If two segments are nearly tied on this metric, the one with the greater distance is chosen. For the Walker2D task, the goal is to move the bipedal robot to the right as far as possible while minimizing energy consumption. Segments where the robot is about to fall or walks abnormally (e.g., using only one leg or slipping) are rated lower, even if the distance covered is longer. If two segments are nearly tied on this metric, the one with the greater distance is chosen. For the \emph{medium-replay} offline dataset, there are 500 queries, while for the \emph{medium-expert} offline dataset, there are 100 queries in total. The segment length for all datasets is $H=100$. 

The Meta-World datasets included script preferences came from \cite{ipl}. First, the datasets included 100 trajectories of expert data for the target task, adding Gaussian noise with a standard deviation of 1.0. Then, the datasets included 100 trajectories of sub-optimal data by running the ground truth policy with noise on different randomizations of the target task, and another 100 trajectories of sub-optimal data by running the ground truth policy of different tasks within the target domain with noise. Finally, the datasets included 100 trajectories generated using uniform random actions. Each Meta-World task dataset contains 200,000 time steps. The preference datasets were constructed by uniformly sampling segments and assigning preference labels based on the total rewards of the segments.

\section{Algorithm Implementations}\label{appsec:implementation}

In this section, we detail the implementations of both HPL and the baseline algorithms used in this paper. 

\subsection{Preference Learning Methods}
\textbf{Markovian Reward (MR). }The MR method optimizes a markovian reward function $r_\psi(s, a)$ using the Bradley-Terry model and the preference dataset $\mathcal{D}_{\rm p}$. The hyper-parameters for MR are listed in Table~\ref{apptab:mr_params}. It is worth noting that we add a final activation layer to the reward network to scale the reward to $[0, 1]$. We find that without such activation, the performance of RL severely deteriorates in some of the Gym MuJoCo tasks. We suspect that this is related to the \textit{survival instinct}~\citep{survival_instinct} in offline RL, i.e. in environments with terminal conditions, negative rewards tend to incline the agent to terminate the trajectory by selecting those dangerous out-of-distribution actions. Based on this observation, we decided to activate the reward values with ReLU for the Hopper and Walker2D tasks while leaving them unchanged for other tasks without environmental terminations. Such activation is shared across MR, PT and HPL. However, one may argue that the activation implicitly imposes an inductive bias on the obtained reward, which may not align with the ground truth. So we also add the reference scores in Section~\ref{sec:main_results} for Gym MoJoCo tasks for comprehensive comparisons. 

\begin{table}[htbp]
    \centering
    \caption{Hyper-parameters for MR.}
    \label{apptab:mr_params}
    \begin{tabular}{c|c}
        \toprule
         hidden dimension for $r_\psi$ & 256 \\
         \# of hidden layers for $r_\psi$ & 2 for Gym MuJoCo tasks and 3 for others \\
         final activation & ReLU for Gym MuJoCo tasks and Identity for others \\
         optimizer & Adam\\
         learning rate & 3e-4 \\
         training steps for $r_\psi$ & 50k \\
         \bottomrule
    \end{tabular}
\end{table}

\textbf{Preference Transformer (PT). }The Preference Transformer employs a causal transformer followed by a bi-directional attention layer to estimate the reward values. By using the causal transformer, the states and actions can attend to historical tokens and thus the reward can utilize the historical information. The final bi-directional attention layer uses the attention scores as the weights of the rewards at each time step. The authors found PT can identify and place more emphasis on those critical states and actions. We also re-implemented the Preference Transformer based on the original Jax implementation provided by the authors, and the hyper-parameters are listed in Table~\ref{apptab:pt_params}. Note that we do not use any validation to select the reward model. 

\begin{table}[htbp]
    \centering
    \caption{Hyper-parameters for PT.}
    \label{apptab:pt_params}
    \begin{tabular}{c|c}
        \toprule
        attention embedding dimension & 256 \\
        \# of attention layers & 3 \\
        \# of attention heads & 1 \\
        dropout rate & 0.1 \\
        final activation & ReLU for Gym MuJoCo tasks and Identity for others \\
        optimizer & Adam\\
        learning rate & 3e-4 \\
        learning rate warm-up steps & 10k \\
        training steps for $r_\psi$ & 100k \\
         \bottomrule
    \end{tabular}
\end{table}

\textbf{Inverse Preference Learning (IPL). }IPL removes the need for learning a reward model, by expressing the reward using the value functions $Q(s, a)$ and $V(s)$ of the RL agent:
\begin{equation}\label{appeq:ipl}
    r(s, a) = Q(s, a) - \gamma \mathbb{E}_{s'\sim p(s'|s, a)}\left[V(s')\right].
\end{equation}
By substituting Eq~\eqref{appeq:ipl} into Eq~\eqref{eq:bt_model}, the loss objective $\mathcal{L}_{\rm MR}$ provides guidance to increase the Q values of preferred states and actions. We also re-implemented IPL in this paper, and keep the hyper-parameters of IPL the same as the ones used in the original paper. 

\textbf{Hindsight Preference Learning (HPL). }The key components of HPL are the conditional reward model $r_\psi$ and the VAE. We list the hyper-parameters of these modules in Table~\ref{apptab:hpl_params}. The hyper-parameters are kept the same as listed in Table~\ref{apptab:hpl_params} unless otherwise noted. 

\begin{table}[htbp]
    \centering
    \caption{Hyper-parameters for HPL.}
    \label{apptab:hpl_params}
    \begin{tabular}{c|c|c}
        \toprule
        \multirow{12}{*}{VAE} & attention embedding dim of encoder $q_\theta$ & 256 \\
        & \# of attention layers & 3 \\
        & \# of attention heads & 1 \\
        & dropout rate & 0.1 \\
        & hidden dim for decoder $p_\theta$& 256 for MuJoCo/Adroit, 512 for Meta-World \\
        & \# of hidden layers for decoder $p_\theta$& 256 \\
        & hidden dim for prior $f_\theta$ & 256 for MuJoCo/Adroit, 512 for Meta-World \\
        & \# of hidden layers for prior $f_\theta$ & 2 \\
        & dimension of embedding $z$ & 128 \\
        & posterior/prior distribution & \revise{diagonal Gaussian}{categorical distribution} \\
        & learning rate & 3e-4 \\
        & training steps & 250k \\
        & KL loss coefficient & \revise{1.0}{0.1} \\
        & encoded future segment length $k$ & 5 \\
        \midrule
        \multirow{6}{*}{$r_\psi$} & hidden dims of $r_\psi$ & 256\\
        & \# of hidden layers for $r_\psi$ & 3 \\
        &final activation & ReLU for MuJoCo, Identity for others \\
        &optimizer & Adam\\
        &learning rate & 3e-4 \\
        & training steps & 100k \\
        & marginalization samples $N$ & 20 \\
         \bottomrule
    \end{tabular}
\end{table}

\subsection{RL Policy Optimization}
For those methods that follow the two-phase paradigm as we discussed in Section~\ref{sec:preliminary_oprl}, we use Implicit Q-Learning (IQL)~\citep{kostrikov2022offline} for policy optimization with the learned reward model. The hyper-parameters for IQL are shared across various reward learning methods for fair comparisons. We list the hyper-parameters in Table~\ref{apptab:iql_params}. 
\begin{table}[htbp]
    \centering
    \caption{Hyper-parameters for IQL.}
    \label{apptab:iql_params}
    \begin{tabular}{c|c}
        \toprule
        expectile $\tau$ & 0.7 for MuJoCo, 0.75 for others\\
        inverse temperature & 0.333 \\
        clipping threshold for $\exp A$ & 100.0 \\
        discounting factor & 0.99 \\
        soft update for target networks & 0.005 \\
        policy network & $\texttt{MLP}(\text{dim}(\mathcal{S}), 256, 256, 256, 2*\text{dim}(\mathcal{A}))$ \\
        policy distribution & tanh-squashed diagonal Gaussian \\
        critic network & $\texttt{MLP}(\text{dim}(\mathcal{S})+\text{dim}(\mathcal{A}), 256, 256, 256, 1)$ \\
        value network & $\texttt{MLP}(\text{dim}(\mathcal{S}), 256, 256, 256, 1)$\\
        optimizers for all networks & Adam \\
        learning rates for all networks & 3e-4 \\
        training steps & 500k \\
         \bottomrule
    \end{tabular}
\end{table}

\subsection{Supervised Fine-tuning}
Finally, we provide details about the implementations of the Supervised Fine-tuning (SFT) method used in the experiment section. For SFT, we use the preferred trajectory segments to perform behavior cloning. The behavior cloning process maximizes the log probability of the policy selecting the preferred segment. Thus, SFT methods fail to leverage the vast offline datasets, which is identified as a key advantage of offline PbRL methods. The hyper-parameters of SFT can be found in Table~\ref{apptab:sft_params}. 

\begin{table}[htbp]
    \centering
    \caption{Hyper-parameters for SFT.}
    \label{apptab:sft_params}
    \begin{tabular}{c|c}
        \toprule
        policy network & $\texttt{MLP}(\text{dim}(\mathcal{S}), 256, 256, 256, \text{dim}(\mathcal{A}))$ \\
        policy distribution & deterministic (Dirac distribution) \\
        optimizer & Adam \\
        learning rates & 3e-4 \\
        training steps & 500k \\
         \bottomrule
    \end{tabular}
\end{table}

\section{Experimental Setups}
In this section, we provide additional details for the main results in Section~\ref{sec:experiments}. 

\textbf{Benchmark results (Table~\ref{tab:main_result}). }We use the full amount of preference datasets and unlabeled datasets as detailed in Section~\ref{appsec:dataset} for MuJoCo tasks and Adroit tasks. For Meta-World tasks, we take the first 500 queries as the preference dataset $\mathcal{D}_{\rm p}$ and the first 5000 queries as the unlabeled dataset $\mathcal{D}_{\rm u}$. The $\mathcal{D}_{\rm p}$ and $\mathcal{D}_{\rm u}$ matches each other in terms of the data source. 

\textbf{Results of the mismatched tasks (Figure~\ref{fig:exp_mismatch}). }We created a series of tasks by cross-over the preference datasets and the unlabeled datasets, as detailed in the main text. Besides, we use the full amount of the datasets, without further selection. 

\textbf{Scaling trends with varied sizes of $\mathcal{D}_{\rm u}$ (Figure~\ref{fig:exp_uscale}). }In this set of experiments, we keep the setups and the hyper-parameters exactly the same as in Table~\ref{tab:main_result}, except for the sizes of the unlabeled dataset. Specifically, we select the first 1k, 2.5k, 5k, 7.5k and 10k trajectories from the dataset (which corresponds to 10\%, 25\%, 50\%, 75\%, 100\% of the total capacity of $\mathcal{D}_{\rm u}$).

\textbf{Scaling trends with varied sizes of $\mathcal{D}_{\rm p}$ (Figure~\ref{fig:exp_pscale}). }In this set of experiments, we keep the setups and the hyper-parameters exactly the same as in Table~\ref{tab:main_result}, except for the sizes of the preference dataset. We select the first 100, 200, 300, 400, 500, and 1000 queries from the dataset, as depicted in the figure. 

\section{Disclosure of computational resources and efficiency}

\begin{figure*}[htbp]
    \centering
    \includegraphics[width=0.7\linewidth]{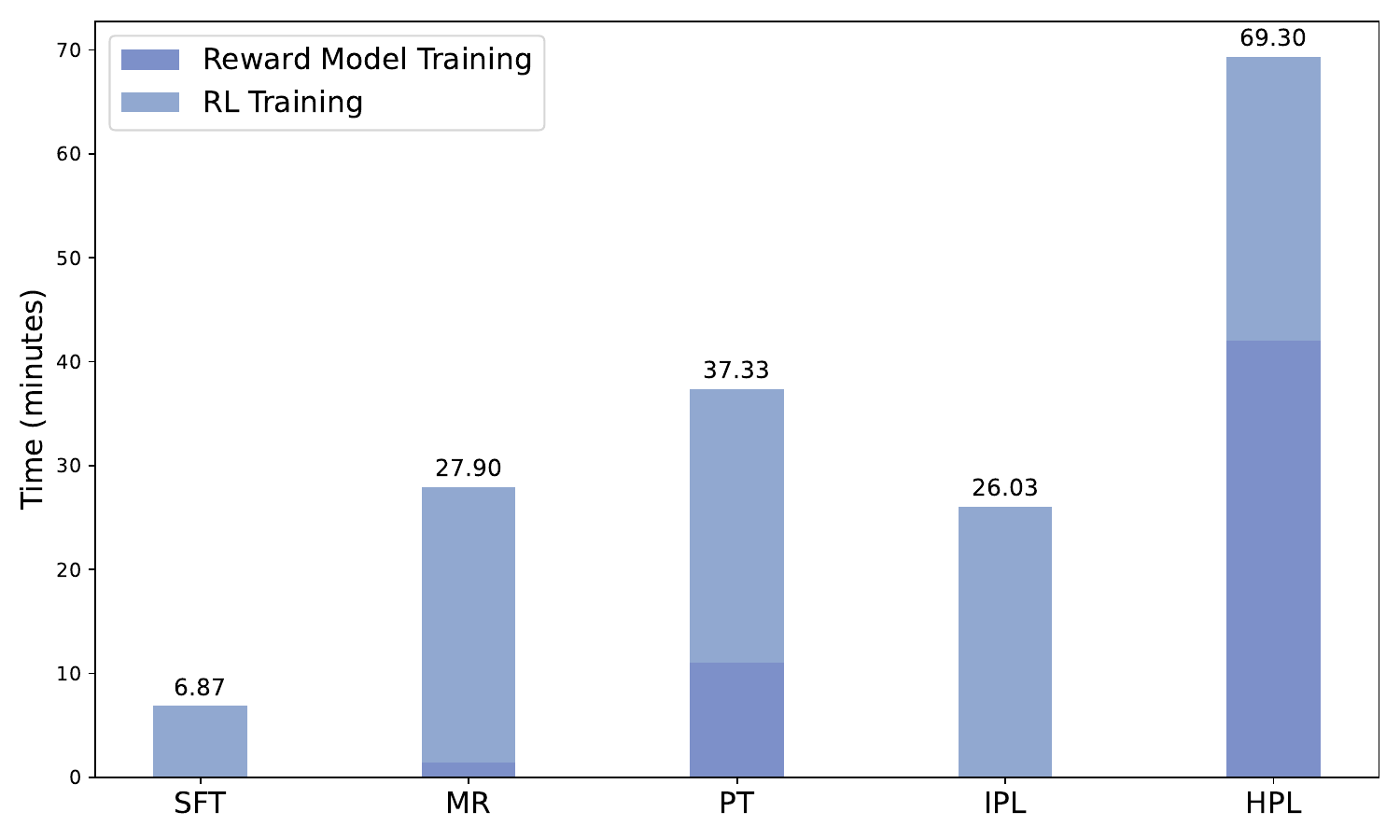}
    \caption{The running time of HPL and each baseline methods, using button-press from Meta-World as an example.}
    \label{fig:algo_time}
\end{figure*}

Throughout the experiments, we evaluate HPL as well as other baseline methods with workstations equipped with NVIDIA RTX 4090 cards. The running time of each method for the button-press task in the Meta-World environment is presented in Figure~\ref{fig:algo_time}.

\section{Supplimentary Experiment Results}
Due to the limited space of the main text, we present additional supplementary results in this section. 

\revise{}{
\subsection{Benchmark Results of AWAC Variants}\label{appsec:awac}
\input{tables/awac_result}

In Section~\ref{sec:main_results}, the results are obtained by using IQL~\citep{kostrikov2022offline} as the policy optimization algorithm. However, given that IQL relies on expectile regression rather than the policy for bootstrapping, it may not fully reveal the potential shortcomings of the learned rewards. Additionally, the choice of expectile could significantly affect the outcomes. In this section, we instead implement AWAC, another offline reinforcement learning (RL) algorithm that integrates policy into bootstrapping, to both HPL and the baseline algorithms. This approach aims to provide a more thorough assessment of reward quality.

The results are listed in Table~\ref{tab:awac_result}. Similar to HPL-IQL, HPL-AWAC demonstrates stable and consistent advantages over the baseline methods in most of the tasks. Overall, HPL-AWAC achieves the best performance on average across these three task suites. 

\subsection{Ablation on the Effect of Ensembling Reward Models}
\begin{figure*}[htbp]
    \centering
    \includegraphics[width=1.0\linewidth]{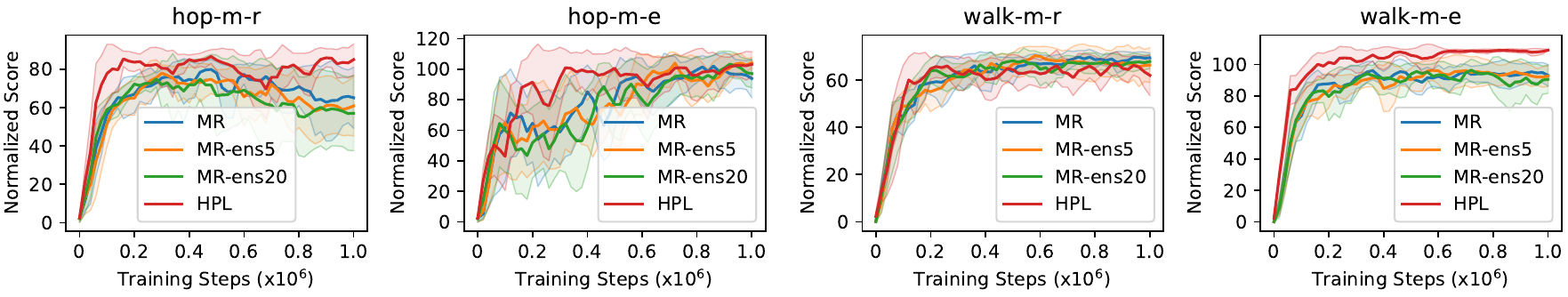}
    \caption{Learning curve of HPL and MR variants with ensemble reward networks. We report the average and the standard deviation of the performances across 10 evaluation episodes and 5 seeds. }
    \label{fig:app_ens}
\end{figure*}

One could argue that the success of HPL can be attributed to the marginalization step (Eq~\eqref{eq:marginalization}), which implicitly ensembles reward models to yield improved rewards for subsequent RL optimization. As revealed by previous literature, reward model ensembling does bring benefit to the credit assignment by characterizing the aleatoric uncertainty and thus facilitating active knowledge acquisition \citep{liang2022reward} or promoting pessimism \citep{costereward}. In this section, we ablate the effect of reward model ensembling by applying the ensembling trick to the MR method. Specifically, we set the number of reward ensembles to 5 and 20 respectively, and keep other configurations the same as the experiments in Section~\ref{sec:main_results}. These reward models are trained with the same dataset, differing only in their initializations. When labeling the dataset, we take the average of the outputs of the ensembles as the reward. Note that this can be considered as an implementation of OPRL, which also employs the ensembling technique.  

\begin{figure*}[htbp]
    \centering
    \includegraphics[width=1.0\linewidth]{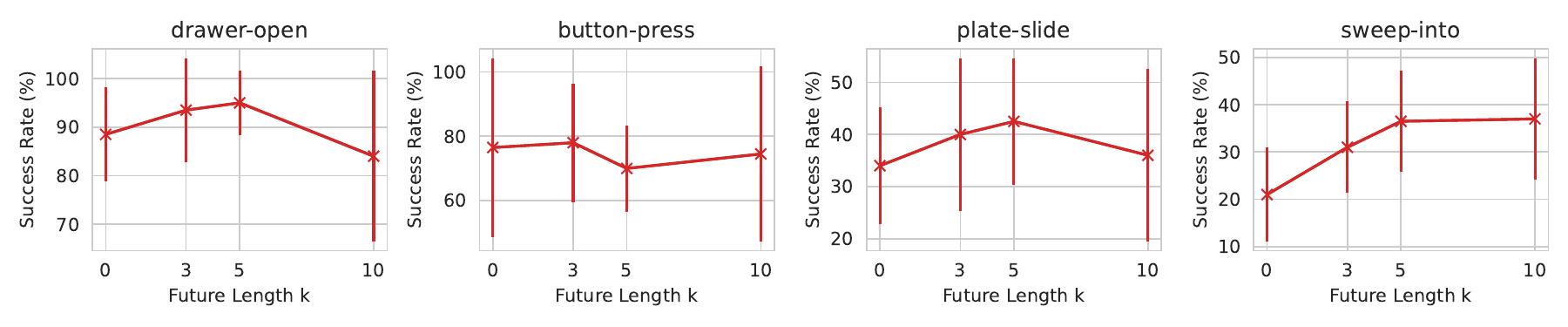}
    \caption{Success rate of HPL with various encoded future length $k$. We report the average and the standard deviation of the performances across 20 evaluation episodes and 10 seeds. }
    \label{fig:app_future_len}
\end{figure*}

As witnessed in Figure~\ref{fig:app_ens}, despite the ensembling technique, the performance of MR still falls behind HPL. This justifies that, naively ensembling reward models which are trained via different initialization does not benefit the credit assignment. On the other hand, HPL ensembles reward values conditioned on different future outcomes based on their prior probabilities, producing more reliable and advantageous rewards. 

\subsection{Ablation on Future Length $k$}
The parameter k controls the lengths of future segments encoded into the embedding. At the extreme of $k\to 0$, HPL theoretically degenerates to MR as the conditional reward $r_\psi$ contains no information about the future. 

Figure~\ref{fig:app_future_len} illustrates the performance of HPL across various values of $k$ for all tasks in Meta-World. As $k$ increases from zero, the performance generally improves, supporting the efficacy of future conditioning. However, beyond a certain threshold, further increases in $k$ lead to performance declines and fluctuations. This phenomenon may be attributed to the incapability of modeling excessively long trajectory segments with the VAE structure. 

\begin{figure*}[htbp]
    \centering
    \includegraphics[width=1.0\linewidth]{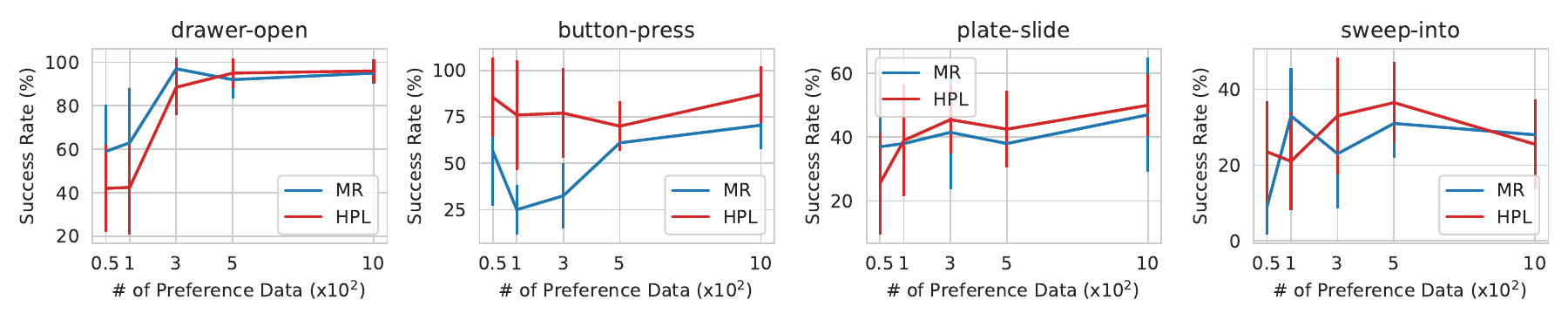}
    \caption{Success rate of HPL and MR with various sizes of $\mathcal{D}_{\rm p}$. We report the average and the standard deviation of the performances across 20 evaluation episodes and 10 seeds. }
    \label{fig:app_pscale}
\end{figure*}

\begin{figure*}[htbp]
    \centering
    \includegraphics[width=1.0\linewidth]{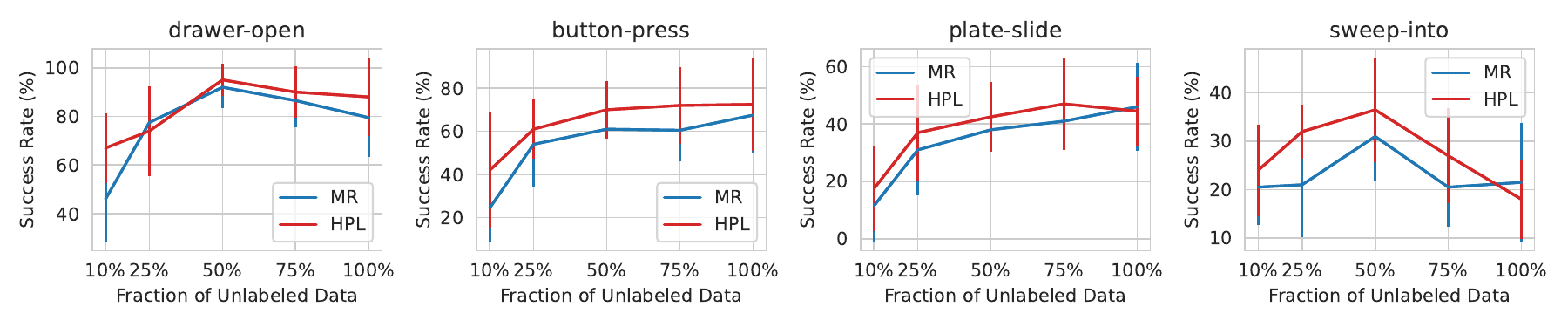}
    \caption{Success rate of HPL and MR with various sizes of $\mathcal{D}_{\rm u}$. We report the average and the standard deviation of the performances across 20 evaluation episodes and 10 seeds. }
    \label{fig:app_uscale}
\end{figure*}

\subsection{Scaling with Dataset Sizes}

In this section, we investigate the performance of HPL as well as MR with various dataset sizes. We vary the sizes of the preference dataset $\mathcal{D}_{\rm p}$ and the unlabeled dataset $\mathcal{D}_{\rm u}$, and plot the curve of the performances in Figure~\ref{fig:app_pscale} and Figure~\ref{fig:app_uscale}, respectively. While both methods exhibit an upward trend in success rates as the dataset sizes $|\mathcal{D}_{\rm p}|$ and $|\mathcal{D}_{\rm u}|$ grow, HPL outperforms MR in almost every configuration. These experiments collectively confirm the superiority and scalability of HPL. 
}

%% file: tables/awac_result.tex
\begin{table*}[htbp]
\caption{Normalized averaged score for AWAC variants of HPL and baseline algorithms. In the table, we use the same abbreviations for tasks as in Table~\ref{tab:main_result}. We report the average and the standard deviation of the performances across 10 evaluation episodes and 5 seeds, and bold the values that are within $95\%$ of the top-performing methods among all methods except for the \textit{Oracle}.}
\label{tab:awac_result}
\centering
\setlength{\tabcolsep}{1.8mm}{}
\begin{tabular}{ccccccc}
\toprule[1.5pt]
Dataset & \textbf{Oracle} & \textbf{SFT} & \textbf{MR-AWAC} & \textbf{PT-AWAC} & \textbf{IPL-AWAC} & \textbf{HPL-AWAC}\\
\midrule
hop-m-r & $97.4$ & $22.2$ & $31.2_{\pm 0.2}$ & $68.7_{\pm 18.3}$ & $69.8_{\pm 13.6}$ & $\bm{94.6_{\pm 3.1}}$\\
hop-m-e & $107.4$ & $5.2$ & $70.9_{\pm 34.5}$ & $\bm{93.3_{\pm 13.9}}$ & $55.3_{\pm 17.2}$ & $\bm{98.0_{\pm 15.9}}$\\
walk-m-r & $82.2$ & $9.0$ & $63.1_{\pm 9.1}$ & $\bm{77.6_{\pm 5.4}}$ & $8.9_{\pm 11.0}$ & $71.2_{\pm 4.0}$\\
walk-m-e & $111.7$ & $0.4$ & $\bm{91.7_{\pm 38.7}}$ & $76.7_{\pm 47.4}$ & $46.3_{\pm 52.8}$ & $\bm{95.1_{\pm 8.5}}$\\
\midrule
Gym average & $99.7$ & $9.2$ & $64.2$ & $79.1$ & $45.1$ & $\bm{89.7}$\\
\midrule
\midrule
pen-h & $78.5$ & $35.4$ & $10.5_{\pm 9.9}$ & $0.0_{\pm 3.9}$ & $11.7_{\pm 8.2}$ & $\bm{44.7_{\pm 27.6}}$\\
pen-c & $83.4$ & $31.1$ & $8.9_{\pm 11.8}$ & $12.9_{\pm 14.3}$ & $13.0_{\pm 17.6}$ & $\bm{36.4_{\pm 27.8}}$\\
ham-h & $1.8$ & $0.3$ & $0.3_{\pm 0.5}$ & $0.0_{\pm 0.1}$ & $0.0_{\pm 0.2}$ & $\bm{4.7_{\pm 5.9}}$\\
ham-c & $1.5$ & $\bm{2.6}$ & $0.1_{\pm 0.2}$ & $0.1_{\pm 0.1}$ & $0.1_{\pm 0.1}$ & $0.2_{\pm 0.0}$\\
\midrule
Adroit average & $41.3$ & $17.4$ & $5.0$ & $3.3$ & $6.2$ & $\bm{21.5}$\\
\midrule
\midrule
drawer-open & - & $0.42$ & $0.77_{\pm 0.28}$ & $0.56_{\pm 0.29}$ & $0.58_{\pm 0.19}$ & $\bm{0.89_{\pm 0.07}}$\\
button-press & - & $0.26$ & $\bm{0.78_{\pm 0.14}}$ & $0.67_{\pm 0.23}$ & $0.66_{\pm 0.26}$ & $0.69_{\pm 0.12}$\\
plate-slide & - & $0.26$ & $0.35_{\pm 0.25}$ & $0.07_{\pm 0.10}$ & $\bm{0.52_{\pm 0.18}}$ & $0.47_{\pm 0.21}$\\
sweep-into & - & $0.24$ & $0.30_{\pm 0.19}$ & $0.10_{\pm 0.12}$ & $0.24_{\pm 0.11}$ & $\bm{0.49_{\pm 0.09}}$\\
\midrule
Meta-World average & - & $0.30$ & $0.55$ & $0.35$ & $0.50$ & $\bm{0.64}$\\

\bottomrule[1.5pt]
\end{tabular}
\end{table*}

%% file: tex/checklist.tex
\section*{NeurIPS Paper Checklist}

\begin{enumerate}

\item {\bf Claims}
    \item[] Question: Do the main claims made in the abstract and introduction accurately reflect the paper's contributions and scope?
    \item[] Answer: \answerYes{} 
    \item[] Justification: The main claims in the abstract include: 1) the Markovian reward assumption is flawed (from Section 3.1); 2) our proposed method accounts for the influence of the future (Section 3.2-3.4); 3) HPL can deliver robust and advantageous rewards (from Section 5.1 and 5.2). 

\item {\bf Limitations}
    \item[] Question: Does the paper discuss the limitations of the work performed by the authors?
    \item[] Answer: \answerYes{} 
    \item[] Justification: We discuss the limitations of HPL in the conclusion section.

\item {\bf Theory Assumptions and Proofs}
    \item[] Question: For each theoretical result, does the paper provide the full set of assumptions and a complete (and correct) proof?
    \item[] Answer: \answerNA{} 
    \item[] Justification: This paper does not include theoretical results.

    \item {\bf Experimental Result Reproducibility}
    \item[] Question: Does the paper fully disclose all the information needed to reproduce the main experimental results of the paper to the extent that it affects the main claims and/or conclusions of the paper (regardless of whether the code and data are provided or not)?
    \item[] Answer: \answerYes{} 
    \item[] Justification: The authors provide all the necessary information to reproduce the experiment results, including the code, environment setups, algorithm implementations, hyper-parameters.

\item {\bf Open access to data and code}
    \item[] Question: Does the paper provide open access to the data and code, with sufficient instructions to faithfully reproduce the main experimental results, as described in supplemental material?
    \item[] Answer: \answerYes{} 
    \item[] Justification: The paper is accompanied by its experimental code, which provides instructions to set up the environment, download the datasets and reproduce the results. The code has been anonymized as well. 

\item {\bf Experimental Setting/Details}
    \item[] Question: Does the paper specify all the training and test details (e.g., data splits, hyperparameters, how they were chosen, type of optimizer, etc.) necessary to understand the results?
    \item[] Answer: \answerYes{} 
    \item[] Justification: We provide detailed algorithm configurations in Appendix B and additional experiment setups in Appendix C.

\item {\bf Experiment Statistical Significance}
    \item[] Question: Does the paper report error bars suitably and correctly defined or other appropriate information about the statistical significance of the experiments?
    \item[] Answer: \answerYes{} 
    \item[] Justification: All of our empirical evaluations present the statistical significance via reporting the one standard deviation across multiple runs with different random seeds. 

\item {\bf Experiments Compute Resources}
    \item[] Question: For each experiment, does the paper provide sufficient information on the computer resources (type of compute workers, memory, time of execution) needed to reproduce the experiments?
    \item[] Answer: \answerYes{} 
    \item[] Justification: The authors provided such information in Appendix D. 
    
\item {\bf Code Of Ethics}
    \item[] Question: Does the research conducted in the paper conform, in every respect, with the NeurIPS Code of Ethics \url{https://neurips.cc/public/EthicsGuidelines}?
    \item[] Answer: \answerYes{} 
    \item[] Justification: The research in this paper conforms with the Code of Ethics.

\item {\bf Broader Impacts}
    \item[] Question: Does the paper discuss both potential positive societal impacts and negative societal impacts of the work performed?
    \item[] Answer: \answerYes{} 
    \item[] Justification: This paper proposes a novel method to extract robust and advantageous rewards from human preferences, which the authors do not find to be harmful to society. 
    
\item {\bf Safeguards}
    \item[] Question: Does the paper describe safeguards that have been put in place for responsible release of data or models that have a high risk for misuse (e.g., pretrained language models, image generators, or scraped datasets)?
    \item[] Answer: \answerNA{} 
    \item[] Justification: The paper is not releasing any new datasets or models that have a high risk for misuse. 

\item {\bf Licenses for existing assets}
    \item[] Question: Are the creators or original owners of assets (e.g., code, data, models), used in the paper, properly credited and are the license and terms of use explicitly mentioned and properly respected?
    \item[] Answer: \answerYes{} 
    \item[] Justification: We provide the introduction, license as well as download links for all of the datasets in this paper. The original creators of these datasets are also credited. 

\item {\bf New Assets}
    \item[] Question: Are new assets introduced in the paper well documented and is the documentation provided alongside the assets?
    \item[] Answer: \answerYes{} 
    \item[] Justification: This paper is only accompanied by its experimental code, without introducing new assets or models. The code has been anonymized. 

\item {\bf Crowdsourcing and Research with Human Subjects}
    \item[] Question: For crowdsourcing experiments and research with human subjects, does the paper include the full text of instructions given to participants and screenshots, if applicable, as well as details about compensation (if any)? 
    \item[] Answer: \answerNA{} 
    \item[] Justification: This paper does not involve crowdsourcing or research with human subjects. It uses human-annotated datasets released by previous research works in this field. We have included the description of the datasets (e.g. how are crowdsourcing workers instructed to provide annotations) in the appendix. 

\item {\bf Institutional Review Board (IRB) Approvals or Equivalent for Research with Human Subjects}
    \item[] Question: Does the paper describe potential risks incurred by study participants, whether such risks were disclosed to the subjects, and whether Institutional Review Board (IRB) approvals (or an equivalent approval/review based on the requirements of your country or institution) were obtained?
    \item[] Answer: \answerNA{} 
    \item[] Justification: This paper does not involve crowdsourcing or research with human subjects. It uses human-annotated datasets released by previous research works in this field. We have included the description, download methods, and license of the dataset in the appendix and we also credited the creators of these datasets properly in the paper. 

\end{enumerate}